%% file: main.tex
\RequirePackage{fix-cm}
\documentclass[10pt,twocolumn,letterpaper]{article}
\usepackage[pagenumbers]{cvpr}
\input{preamble}

\definecolor{cvprblue}{rgb}{0.21,0.49,0.74}
\usepackage[breaklinks,colorlinks,allcolors=cvprblue]{hyperref}
\hypersetup{
  pdftitle={VGGT-Long++: Optimize Scale, Align Geometry, and Adapt Online for Globally Consistent Long-Sequence 3D Reconstruction},
  pdfauthor={Rui Xu, Kai Deng, Siyuan Gao, Yigong Zhang, Jin Xie},
  pdfsubject={Long-sequence 3D reconstruction},
  pdfkeywords={3D reconstruction, scale optimization, geometric alignment, online adaptation},
  bookmarksopen=true,
  bookmarksnumbered=true
}

\title{VGGT-Long++: Optimize Scale, Align Geometry, and Adapt Online\\
for Globally Consistent Long-Sequence 3D Reconstruction}
\input{authors}

\newcommand{\mainref}[1]{\ref{#1}}
\newcommand{\mainequation}[1]{(\mainref{#1})}

\begin{document}
\raggedbottom
\etocdepthtag.toc{main}
\setcounter{footnote}{1}
\maketitle
\input{fig/teaser}
\begingroup
\widowpenalty=10000
\input{sec/0_abstract}
\endgroup
\input{sec/1_intro}
\input{sec/2_related}
\input{sec/3_method}
\input{sec/4_experiments}
\input{sec/5_conclusion}

{\small
  \bibliographystyle{ieeenat_fullname}
  \bibliography{main}
}

\input{supplementary}

\end{document}

%% file: preamble.tex
\usepackage{cuted}
\usepackage{booktabs,colortbl,multirow,bigstrut,tabularx,pifont}
\usepackage{caption}
\usepackage{microtype}
\usepackage{stfloats,needspace}
\usepackage{amsmath,amssymb}
\usepackage{algorithm,algpseudocode,placeins}
\usepackage{etoc}

\newcommand{\cmark}{\raisebox{0.3pt}{\fontsize{7.5}{9}\selectfont\ding{51}}}
\newcommand{\xmark}{\raisebox{0.3pt}{\fontsize{7.5}{9}\selectfont\ding{55}}}
\algrenewcommand{\algorithmicrequire}{\textbf{Input:}}
\algrenewcommand{\algorithmicensure}{\textbf{Output:}}

\makeatletter
\newcommand{\preservepdfgroup}{%
  \let\Gread@transgrouptrue\Gread@transgroupfalse}
\renewcommand*\l@section[2]{%
  \addvspace{0.4\baselineskip}%
  \@dottedtocline{1}{0pt}{1.75em}{\bfseries #1}{\bfseries #2}}
\renewcommand*\l@subsection{\@dottedtocline{2}{0.8em}{2.5em}}
\makeatother

%% file: authors.tex
\author{
  Rui Xu\quad Kai Deng\quad Siyuan Gao\quad Yigong Zhang\quad Jin Xie\thanks{Corresponding author.}\\
  School of Intelligence Science and Technology, Nanjing University\\
  {\tt\small rui.xu@smail.nju.edu.cn\quad csjxie@nju.edu.cn}
}

%% file: fig/teaser.tex
\setlength{\stripsep}{4pt plus 1pt minus 1pt}
\begin{strip}
\centering
{\preservepdfgroup
\includegraphics[width=0.97\linewidth]{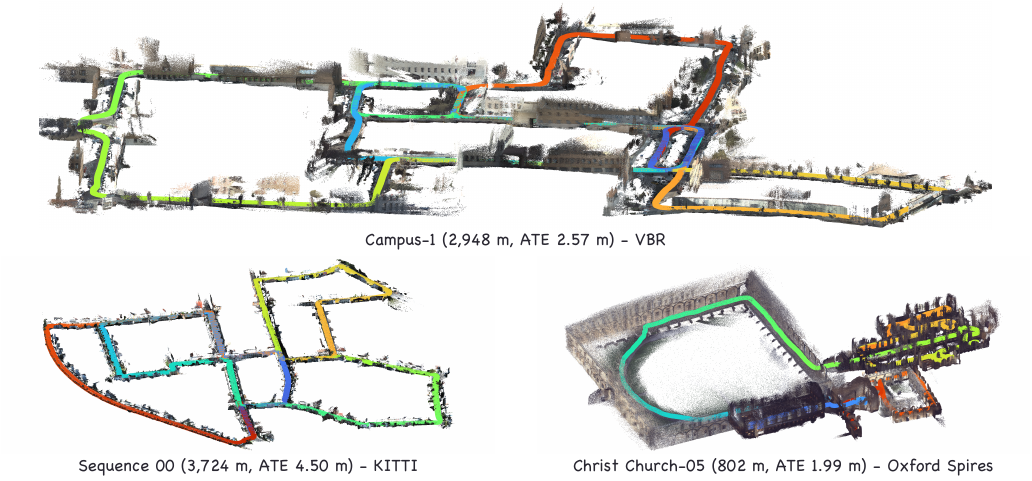}}
\captionsetup{hypcap=false}
\captionof{figure}{\textbf{Long-sequence 3D reconstruction results.} Complete reconstructions from VBR, KITTI, and Oxford Spires are shown, with camera frusta color-coded by time.}
\captionsetup{hypcap=true}
\label{fig:teaser}%
\end{strip}

%% file: sec/0_abstract.tex
\begin{abstract}
Extending feed-forward foundation models to long-sequence 3D reconstruction typically relies on registering adjacent overlapping chunks reconstructed independently. However, the same images may yield inconsistent geometric predictions across chunks, causing drift and degrading global reconstruction quality during sequential registration.
VGGT-Long mitigates this issue through loop constraints and global $\mathrm{Sim}(3)$ optimization.
Yet adjacent registration constraints still rely on local overlaps, while joint $\mathrm{Sim}(3)$ optimization may allow scale and pose errors to compensate for one another.
A single similarity transformation also cannot fully resolve geometric discrepancies between chunks.
Motivated by these observations, we propose \textbf{VGGT-Long++}, which extends this framework toward globally consistent long-sequence 3D reconstruction. Specifically,
scale-wise graph optimization exploits long-range scale constraints from jointly reconstructed keyframe chunks to refine the registration chain of local chunks.
Depth and affine alignment corrects residual geometric discrepancies across registered chunks through layer-wise depth adjustments and chunk-wise affine transformations.
Camera-guided online adaptation further combines consistency and epipolar supervision to progressively improve the accuracy and cross-chunk consistency of subsequent predictions. Without additional offline training, VGGT-Long++ achieves state-of-the-art trajectory accuracy using uncalibrated monocular RGB, with average ATE RMSEs of 10.55, 1.41, and 3.59\,m on KITTI, Oxford Spires, and VBR, respectively.
Further evaluations demonstrate improved dense reconstruction quality and efficient streaming execution.
\textbf{Project page:}~\url{https://ruixu15.github.io/vggt-long-plus-plus/}.
\end{abstract}

%% file: sec/1_intro.tex
\section{Introduction}
\label{sec:intro}

Reconstructing large-scale 3D scenes from long monocular image sequences is essential for autonomous navigation, robotics, and spatial understanding~\cite{schoenberger2016sfm,campos2021orbslam3}.
Recent feed-forward 3D foundation models, such as DUSt3R~\cite{wang2024dust3r}, VGGT~\cite{wang2025vggt}, and $\pi^3$~\cite{wang2025pi3}, directly predict dense  geometry from uncalibrated images.
However, their computational demands make inference over extended sequences impractical~\cite{ding2026laser}.
A common solution is to independently reconstruct overlapping local chunks and combine their predictions through chunk-wise registration~\cite{deng2026vggtlong,maggio2025vggtslam}.

Despite overlapping inputs, these models still produce geometrically inconsistent predictions across chunks~\cite{ding2026laser,maggio2026vggtslam2,zhou2026online3r}.
Such inconsistencies may perturb adjacent registrations and
accumulate along the registration chain of local chunks,
causing drift and degrading global reconstruction quality~\cite{jiang2026geoweaver,zhang2026vggtalign}.
To mitigate this issue, VGGT-Long~\cite{deng2026vggtlong} performs confidence-weighted robust registration between adjacent chunks and applies global $\mathrm{Sim}(3)$ optimization with loop constraints derived from jointly reconstructed revisited views.

However, adjacent registration constraints in VGGT-Long are still estimated from 3D point correspondences in local overlaps, and the local registration chain connects distant chunks by composing these pairwise estimates.
Jointly optimizing scale and pose may also allow errors in one to compensate for errors in the other. Moreover, a single similarity transformation per chunk cannot fully resolve shape discrepancies arising from differences in estimated
depth and camera ray directions across chunks.

Motivated by these observations, we propose \textbf{VGGT-Long++},
which extends the VGGT-Long framework toward globally consistent
long-sequence 3D reconstruction.
We first introduce jointly reconstructed keyframe chunks to
provide long-range scale constraints, enabling
\emph{scale-wise graph optimization} to refine the
registration chain of local chunks.
We then develop \emph{depth and affine alignment} to correct
residual geometric discrepancies that remain
after chunk-wise registration.
We further introduce \emph{camera-guided online adaptation}
to improve subsequent model predictions through camera
consistency and epipolar supervision.
Together, these designs reduce the propagation of prediction
errors from local chunks into the global reconstruction.

We evaluate VGGT-Long++ on challenging long-sequence benchmarks, including KITTI~\cite{geiger2012kitti}, Oxford Spires~\cite{tao2025spires}, and VBR~\cite{brizi2024vbr}, with representative reconstructions in Fig.~\ref{fig:teaser}. VGGT-Long++ achieves state-of-the-art trajectory accuracy using
uncalibrated monocular RGB without additional offline training.
Quantitative and qualitative evaluations further demonstrate
improved dense reconstruction quality. The complete system achieves an average streaming throughput of 28.86 FPS across KITTI sequences on a single NVIDIA RTX 4090 GPU.

Our main contributions are summarized as follows.
\begin{itemize}
\item We introduce scale-wise graph optimization, which  integrates long-range scale constraints from joint inference within keyframe chunks to refine the registration chain of local chunks.

\item {\interlinepenalty=10000 We develop depth and affine alignment to correct residual geometric  discrepancies beyond a single similarity transformation.\par}

\Needspace{3\baselineskip}
\item We propose camera-guided online adaptation that combines consistency and epipolar supervision to improve the accuracy and cross-chunk consistency of subsequent model predictions.

\item Extensive experiments demonstrate state-of-the-art trajectory accuracy using uncalibrated monocular RGB, together with improved dense reconstruction quality and efficient streaming execution.
\end{itemize}

%% file: sec/2_related.tex
\section{Related work}
\label{sec:related}

\paragraph{Feed-forward 3D reconstruction.}

Classical structure-from-motion (SfM)~\cite{schoenberger2016sfm}
and visual simultaneous localization and mapping
(SLAM)~\cite{campos2021orbslam3} recover scene structure and camera
motion through feature correspondences and geometric optimization.
Feed-forward 3D foundation models directly predict scene geometry
from images.
DUSt3R~\cite{wang2024dust3r} formulates pairwise reconstruction as
pointmap regression, while MASt3R~\cite{leroy2024mast3r} augments this
representation with learned features for dense matching.
Fast3R~\cite{yang2025fast3r} and MUSt3R~\cite{cabon2025must3r}
extend reconstruction to multiple views in a shared coordinate frame.
VGGT~\cite{wang2025vggt} further integrates camera estimation
with dense reconstruction, while $\pi^3$~\cite{wang2025pi3}
adopts a permutation-equivariant architecture to predict camera
poses and local pointmaps.
Depth Anything 3~\cite{lin2025depthanything3} recovers multi-view
geometry through depth and ray prediction, and
MapAnything~\cite{keetha2026mapanything} integrates images with
optional geometric inputs for metric reconstruction.

\input{fig/pipeline}
\paragraph{Long-sequence 3D reconstruction.}
Extending feed-forward 3D reconstruction to long sequences requires
integrating geometric information over time.
VGGT-Long~\cite{deng2026vggtlong} aligns overlapping chunks and
incorporates loop constraints into global $\mathrm{Sim}(3)$ optimization.
SLAM3R~\cite{liu2025slam3r} learns local-to-global pointmap registration,
while ABot-Recon~\cite{amap2026abot} composes locally estimated relative poses.
Other approaches retain scene information through
memory~\cite{wang2025spann3r,wang2025continuous,chen2025long3r,liu2026mem3r},
cached attention~\cite{zhuo2025streamvggt,yuan2026infinitevggt},
or adaptive updates~\cite{chen2026ttt3r,xie2026scal3r}.
LingBot-Map~\cite{chen2026lingbot} organizes attention over reference
views and trajectory memory, while
HorizonStream~\cite{cheng2026horizonstream} propagates geometry
across temporal scales.
LoGeR~\cite{zhang2026loger} combines fast weights with sliding-window
attention, and CLoSeR~\cite{li2026closer} extends it with loop
constraints and pose graph optimization.
VGGT-Bridge~\cite{choi2026vggtbridge} and
AMB3R-SLAM~\cite{wang2026amb3rslam} introduce skip edges and
hierarchical constraints, respectively.
For geometric refinement, SURE-Map~\cite{liu2026suremap} combines uncertainty-weighted
translation refinement with scale recalibration of recent
trajectory segments, while LASER~\cite{ding2026laser} aligns depth layers
across windows.
VGGT-SLAM 2.0~\cite{maggio2026vggtslam2} enforces pose and calibration
consistency, and GeoWeaver~\cite{jiang2026geoweaver} jointly refines
poses, depth, and intrinsics.
In contrast, VGGT-Long++ refines the registration chain of local chunks
through scale-wise graph optimization and corrects residual
geometric discrepancies through depth and affine alignment.

\paragraph{Test-time adaptation for 3D reconstruction.}
Test-time training~\cite{sun2020ttt}, Tent~\cite{wang2021tent},
and CoTTA~\cite{wang2022cotta} adapt pretrained models
using unlabeled test data.
View synthesis~\cite{zhou2017sfmlearner},
3D alignment~\cite{mahjourian2018geometric}, and cross-frame depth
consistency~\cite{bian2019scaleconsistent} provide geometric
self-supervision for depth and motion estimation.
Consistent Video Depth Estimation~\cite{luo2020consistent} further
uses geometric consistency to fine-tune a pretrained depth network
on individual videos.
More recently, Online3R~\cite{zhou2026online3r} learns visual prompts
from temporally fused pointmaps and consistency across reference views.
VGGT-Align~\cite{zhang2026vggtalign} adapts normalization parameters
using photometric and geometric supervision, while
Self-Geometry~\cite{youn2026selfgeometry} uses image correspondences
for parameter-efficient adaptation.
In contrast, VGGT-Long++ exploits camera consistency on overlapping
images across adjacent local chunks for camera-guided online adaptation.

%% file: fig/pipeline.tex
\begin{figure*}[t]
\centering
\includegraphics[width=0.98\textwidth]{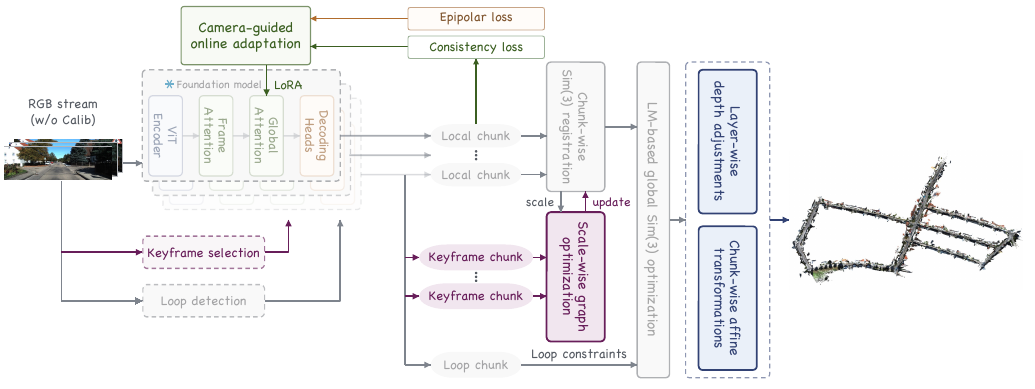}
\caption{\textbf{Overview of VGGT-Long++.}
Scale-wise graph optimization refines the registration chain of local chunks using keyframe chunks.
Depth and affine alignment corrects residual geometric discrepancies across registered chunks.
Camera-guided online adaptation improves subsequent predictions.
The snowflake denotes frozen pretrained weights.}
\label{fig:pipeline}
\end{figure*}

%% file: sec/3_method.tex
\section{Method}
\label{sec:method}

\subsection{Overview}
\label{sec:overview}

\begingroup
\tolerance=1000
\looseness=-1
As illustrated in Fig.~\ref{fig:pipeline}, VGGT-Long++ follows a streaming pipeline that reconstructs local chunks
as images arrive, with alignment proceeding asynchronously.
Scale-wise graph optimization refines the registration chain of local chunks
using jointly reconstructed \emph{keyframe chunks}.
The updated adjacent
constraints are then combined with loop constraints for global $\mathrm{Sim}(3)$ optimization.
Depth and affine alignment corrects residual geometric discrepancies,
while camera-guided online adaptation updates the model to improve subsequent predictions.
\par\endgroup

\subsection{Revisiting VGGT-Long}

Given an incoming monocular image sequence
$\mathcal I=\{I_t\}_{t=0}^{N}$,
VGGT-Long~\cite{deng2026vggtlong} partitions it into overlapping
local chunks $\{\mathcal C_k\}_{k\in\mathcal L}$.
A foundation model independently processes each chunk, producing
a locally consistent geometric representation
$\mathcal G_k=\{\mathbf P_{k,t},d_{k,t},\mathbf T_{k,t},
\mathbf K_{k,t},\chi_{k,t}\mid I_t\in\mathcal C_k\}$.
Here, $\mathbf P_{k,t}$, $d_{k,t}$, $\mathbf K_{k,t}$, and
$\chi_{k,t}$ denote the pointmap in the chunk frame, depth map,
camera intrinsics, and prediction confidence, respectively.
The camera-to-chunk pose
$\mathbf T_{k,t}=(\mathbf R_{k,t},\mathbf c_{k,t})$
consists of rotation $\mathbf R_{k,t}$ and camera center
$\mathbf c_{k,t}$.
Let $S_k\in\mathrm{Sim}(3)$ denote the global transformation
mapping chunk $k$ to the world frame, with $S_0=I$ anchoring
the first chunk.
For adjacent chunks $i$ and $j$, chunk-wise $\mathrm{Sim}(3)$
registration uses the pointmaps of overlapping images to estimate
the relative transformation
$\hat Z_{j\rightarrow i}=(\hat s_{j\rightarrow i},
\hat{\mathbf R}_{j\rightarrow i},\hat{\mathbf t}_{j\rightarrow i})$
from $j$ to $i$.
Sequential composition initializes the global transformations as $S_j=S_i\hat Z_{j\rightarrow i}$.

For an accepted loop between nonadjacent chunks $i$ and $j$,
views around the matched images are jointly reconstructed
as a loop chunk $r$.
Registering $r$ to each local chunk yields $Z_{r\rightarrow i}$
and $Z_{r\rightarrow j}$, defining the loop constraint
$Z_{j\rightarrow i}=Z_{r\rightarrow i}Z_{r\rightarrow j}^{-1}$.
Let $\mathcal E_{\mathrm{adj}}$ and $\mathcal E_{\mathrm{loop}}$
denote the sets of adjacent and loop constraints, respectively,
with initial adjacent measurements given by
$Z_{j\rightarrow i}=\hat Z_{j\rightarrow i}$.
Global $\mathrm{Sim}(3)$ optimization jointly refines
$\{S_i\}_{i\in\mathcal L}$ by solving
\begin{equation}
\label{eq:global_alignment}
\begin{aligned}
\{S_i^\star\}_{i\in\mathcal L}
&=
\underset{\{S_i\}_{i\in\mathcal L},\,S_0=I}{\arg\min}
\sum_{(i,j)\in\mathcal E_{\mathrm{adj}}\cup\mathcal E_{\mathrm{loop}}}
\\
&\quad
\left\|
\operatorname{Log}
\left(Z_{j\rightarrow i}^{-1}S_i^{-1}S_j\right)
\right\|_2^2.
\end{aligned}
\end{equation}
Here, $\operatorname{Log}$ denotes the vector-form logarithm map
of $\mathrm{Sim}(3)$.
The objective is minimized using Levenberg--Marquardt (LM).

Although loop constraints mitigate accumulated drift,
adjacent registration constraints are still estimated from local
overlaps, and joint optimization may allow scale and pose errors
to compensate for one another.
VGGT-Long++ first replaces pointmap registration with separate
scale and pose estimation.
The relative scale is estimated from depth predictions for
overlapping images, while the relative rotation and translation
are obtained by aligning the corresponding camera poses.
\emph{Scale-wise graph optimization} then couples the scales
of temporally distant local chunks through their overlaps with
jointly reconstructed keyframe chunks, thereby refining
the local registration chain.
Yet a single similarity transformation cannot fully resolve
regional depth and calibration discrepancies~\cite{ding2026laser,maggio2026vggtslam2},
motivating \emph{depth and affine alignment}.
\emph{Camera-guided online adaptation} further improves the
accuracy and cross-chunk consistency of subsequent predictions
through supervision from overlapping predictions and image
correspondences.

\subsection{Scale-wise graph optimization}
\label{sec:scale_alignment}

\input{fig/scale_graph}

A jointly reconstructed keyframe chunk provides a shared scale
reference for the local chunks it overlaps.
VGGT-Long++ builds a scale graph using relative scale measurements
from depth predictions for overlapping images.

\paragraph{Keyframe chunks as scale references.}
Starting from the first image, VGGT-Long++ selects keyframes based on
angular parallax relative to the most recently selected keyframe $q$.
Specifically, the angular parallax of a candidate image $I_t$ is computed as
\begin{equation}
	\label{eq:keyframe_selection}
	\alpha_{qt}
	=\underset{u\in\Omega_q}{\operatorname{med}}\,
	\angle\!\left(
	\mathbf p_q(u)-\mathbf c_q,\,
	\mathbf p_q(u)-\mathbf c_t
	\right).
\end{equation}
Here, $\mathbf p_q(u)$ denotes the 3D point back-projected from the
predicted depth of keyframe $q$ at pixel $u$, and $\Omega_q$ denotes
the set of confidence-filtered pixels.
The point and camera centers $\mathbf c_q$ and $\mathbf c_t$ are expressed in the same coordinate frame.

A candidate image $I_t$ is selected when
$\alpha_{qt}\geq\tau_{\mathrm{kf}}$, where
$\tau_{\mathrm{kf}}$ is the angular parallax threshold.
The selected keyframes are grouped into overlapping
keyframe chunks for joint reconstruction.

\paragraph{Scale optimization and registration refinement.}
As illustrated in Fig.~\ref{fig:scale_graph}, VGGT-Long++ constructs a
scale graph $\mathcal G_s=(\mathcal V_s,\mathcal E_s)$, where
$\mathcal V_s=\mathcal L\cup\mathcal K$ contains local and keyframe
chunks, and $\mathcal E_s$ contains local--local, local--keyframe,
and keyframe--keyframe edges established from overlapping images.
Each node $i\in\mathcal V_s$ is assigned a log-scale variable $x_i$,
corresponding to a uniform scale factor $e^{x_i}$ applied to its
3D points and camera centers.
For each edge $(i,j)\in\mathcal E_s$, the relative log-scale
measurement $m_{ij}$ is computed as
\begin{equation}
\label{eq:scale_measurement}
m_{ij}
=
\underset{I_t\in\mathcal O_{ij}}{\operatorname{med}}\,
\underset{u\in\Omega_{ij,t}}{\operatorname{med}}
\bigl(\log d_{i,t}(u)-\log d_{j,t}(u)\bigr).
\end{equation}
Here, $\mathcal O_{ij}$ denotes the set of overlapping images, and
$\Omega_{ij,t}$ contains the pixels that are confidence-filtered in both predictions of image $t$.
The log-scales are then estimated by solving
\begin{equation}
\label{eq:scale_alignment}
\mathbf x^\star
=
\underset{\mathbf x,\;x_0=0}{\arg\min}
\sum_{(i,j)\in\mathcal E_s}
\rho_{\mathrm H}(x_j-x_i-m_{ij}).
\end{equation}
Here, $\rho_{\mathrm H}$ denotes the Huber loss.
The graph is incrementally updated and optimized using iteratively reweighted least squares (IRLS) whenever a new keyframe chunk is integrated.

Using the resulting scales, VGGT-Long++ refines adjacent
registrations by updating their relative scales and recomputing
translations from corresponding camera centers in overlapping images
\begin{equation}
\label{eq:registration_update}
\!s^\star_{j\rightarrow i}=e^{x_j^\star-x_i^\star},\,
\mathbf t^\star_{j\rightarrow i}
=\underset{I_t\in\mathcal O_{ij}}{\operatorname{gmed}}
\bigl(\mathbf c_{i,t}-s^\star_{j\rightarrow i}
\hat{\mathbf R}_{j\rightarrow i}\mathbf c_{j,t}\bigr).
\end{equation}
Here, $\operatorname{gmed}$ denotes the geometric median.
The resulting transformations
$Z_{j\rightarrow i}=(s^\star_{j\rightarrow i},
\hat{\mathbf R}_{j\rightarrow i},\mathbf t^\star_{j\rightarrow i})$
replace $\hat Z_{j\rightarrow i}$ as adjacent constraints in
a Huber-robust form of Eq.~\eqref{eq:global_alignment}.

\subsection{Depth and affine alignment}
\label{sec:geometry_alignment}

Depth determines point positions along camera rays, while camera
intrinsics specify ray directions.
VGGT-Long++ therefore combines layer-wise depth adjustments
with chunk-wise affine transformations, using constraints from
overlapping images across local, keyframe, and loop chunks.

\paragraph{Layer-wise depth adjustments.}
With the chunk scales fixed, each predicted depth map $d_{i,t}$ is represented at a common scale as $\bar d_{i,t}$.
Each depth map is segmented into spatially connected regions with similar depth using graph-based segmentation~\cite{felzenszwalb2004segmentation}, and each region is treated as a depth layer and assigned a log-depth adjustment $\delta_a$.

Let $\mathcal E_d$ denote the set of layer pairs with spatially overlapping masks from the same image across chunks or consecutive frames within a chunk.
For layers $a$ and $b$ predicted for the same image $I_t$ in chunks
$i$ and $j$, the median discrepancy
$\eta^d_{ab}
=\operatorname{med}_{u\in\Omega_{ab}}
(\log\bar d_{i,t}(u)-\log\bar d_{j,t}(u))$
constrains their relative adjustment $\delta_b-\delta_a$.
Here, $\Omega_{ab}$ denotes the intersection of their
confidence-filtered masks.
For pairs from consecutive frames, $\eta^d_{ab}$ is set to zero to encourage temporal consistency. VGGT-Long++ estimates the adjustments by solving
\begin{equation}
\label{eq:depth_alignment}
\min_{\boldsymbol\delta}
\sum_{(a,b)\in\mathcal E_d}
w_{ab}(\delta_b-\delta_a-\eta^d_{ab})^2
+\sum_a w_a\delta_a^2.
\end{equation}
Here, $w_{ab}$ denotes the normalized overlap area, additionally weighted by mask IoU for temporal pairs, while $w_a$ is
the normalized area of layer $a$. The objective is optimized by solving sparse linear systems, with the final term regularizing log-depth adjustments toward zero.

For a pixel $u$ in layer $a$ of image $I_t$ in chunk $i$, the
corresponding point is adjusted along its camera ray as
$\mathbf P^d_{i,t}(u)=\mathbf c_{i,t}
+e^{\delta_a^\star}
(\mathbf P_{i,t}(u)-\mathbf c_{i,t})$.

\paragraph{Chunk-wise affine transformations.}
VGGT-Long++ estimates intrinsic offsets to derive a shared affine
transformation for each local chunk.
The intrinsics of image $I_t$ in chunk $i$ are represented as
\begin{equation}
	\label{eq:intrinsic_features}
	\boldsymbol\kappa_{i,t}
	=
	[
	\log\bar f_x,\,
	\log\bar f_y,\,
	\bar p_x,\,
	\bar p_y
	]^\top.
\end{equation}
Here, $\bar f$ and $\bar p$ denote focal lengths and principal-point
coordinates normalized by the corresponding image dimensions.

Each chunk $i$ is assigned an offset $\mathbf b_i\in\mathbb R^4$,
which is added to $\boldsymbol\kappa_{i,t}$ for all its frames.
Let $\mathcal E_K$ denote the set of chunk pairs with overlapping
images. For each pair $(i,j)\in\mathcal E_K$, the component-wise
median discrepancy
$\boldsymbol\eta^K_{ij}
=\operatorname{med}_{I_t\in\mathcal O_{ij}}
(\boldsymbol\kappa_{i,t}-\boldsymbol\kappa_{j,t})$
constrains the relative offset $\mathbf b_j-\mathbf b_i$.
VGGT-Long++ estimates the offsets by solving
\begin{equation}
\label{eq:intrinsic_alignment}
\min_{\{\mathbf b_i\}}\quad
\sum_{(i,j)\in\mathcal E_K}
\|\mathbf b_j-\mathbf b_i-\boldsymbol\eta^K_{ij}\|_2^2
+\sum_{i\in\mathcal L}\|\mathbf b_i\|_2^2.
\end{equation}
The objective is also optimized by solving sparse linear systems, with the final term regularizing intrinsic offsets of local chunks toward zero.

For each local chunk $i$, the resulting offset $\mathbf b_i^\star$
determines the target intrinsics $\mathbf K'_{i,t}$ from
$\boldsymbol\kappa_{i,t}+\mathbf b_i^\star$.
Averaging the per-camera ray transformations in chunk coordinates
yields a shared affine transformation
\begin{equation}
\label{eq:affine_alignment}
\begin{aligned}
\mathbf A_i
&=
\frac{1}{|\mathcal C_i|}
\sum_{I_t\in\mathcal C_i}
\mathbf R_{i,t}
(\mathbf K'_{i,t})^{-1}
\mathbf K_{i,t}
\mathbf R_{i,t}^{\top},\\
F_i(\mathbf p)
&=
\bar{\mathbf c}_i+
\mathbf A_i(\mathbf p-\bar{\mathbf c}_i).
\end{aligned}
\end{equation}
Here, $\bar{\mathbf c}_i$ denotes the mean camera center in chunk $i$.

Applying $F_i$ followed by $S_i^\star$ to $\mathbf P^d_{i,t}$ yields
the pointmap in the world frame
$\mathbf P^{\mathrm{map}}_{i,t}(u)
=S_i^\star\big(F_i(\mathbf P^d_{i,t}(u))\big)$.
The camera centers undergo the same transformations, and the
corresponding rotations and intrinsics are recovered by RQ
decomposition of the transformed camera projection matrices.

\subsection{Camera-guided online adaptation}
\label{sec:self_tuning}

Adjacent local chunks $i$ and $j$ yield predictions of relative
camera motion and intrinsics for overlapping images.
VGGT-Long++ reuses these predictions to adapt the foundation
model online, combining cross-chunk consistency with epipolar
supervision from image correspondences.

\paragraph{Consistency loss.}
For images $I_t,I_s\in\mathcal O_{ij}$ with $t<s$, the relative
rotation and translation direction from frame $s$ to frame $t$
predicted by chunk $k\in\{i,j\}$ are
\begin{equation}
\label{eq:tuning_relative}
\mathbf Q_{ts}^{k}=\mathbf R_{k,t}^{\top}\mathbf R_{k,s},\quad
\hat{\mathbf v}_{ts}^{k}
=\frac{\mathbf R_{k,t}^{\top}(\mathbf c_{k,s}-\mathbf c_{k,t})}
{\|\mathbf c_{k,s}-\mathbf c_{k,t}\|_2}.
\end{equation}
The consistency loss is defined as
\begin{equation}
	\label{eq:tuning_consistency}
	\begin{aligned}
		\mathcal L_{\mathrm{cons}}
		&=\lambda_R\mathcal L_R+
		\lambda_t\mathcal L_t+
		\lambda_K\mathcal L_K,\\
		\mathcal L_R
		&=
		\left\langle
		d_{\mathrm{SO}(3)}(\mathbf Q_{ts}^{i},\mathbf Q_{ts}^{j})
		\right\rangle_{t,s},\\
		\mathcal L_t
		&=
		\left\langle
		\gamma_{ts}
		\left(1-\hat{\mathbf v}_{ts}^{i}\cdot\hat{\mathbf v}_{ts}^{j}\right)
		\right\rangle_{t,s},\\
		\mathcal L_K
		&=
		\left\langle
		\rho_K(\boldsymbol\kappa_{i,t}-\boldsymbol\kappa_{j,t})
		\right\rangle_t.
	\end{aligned}
\end{equation}
Here, $\langle\cdot\rangle_{t,s}$ and $\langle\cdot\rangle_t$
denote averages over frame pairs and individual frames in
$\mathcal O_{ij}$, respectively. The function $d_{\mathrm{SO}(3)}$ measures geodesic rotation distance,
while $\rho_K$ applies weighted Huber penalties and $\gamma_{ts}$ is computed from
depth-normalized baselines in both predictions.

\paragraph{Epipolar loss.}
Since the two predictions may share errors, VGGT-Long++ complements consistency with epipolar supervision from image correspondences obtained using SIFT~\cite{lowe2004sift} and verified by RANSAC~\cite{fischler1981ransac}.
For a matched image pair $I_t,I_s\in\mathcal C_k$, the predicted
poses and intrinsics determine the fundamental matrix
$\mathbf F_{k,ts}$, which maps points in image $s$ to epipolar
lines in image $t$.
The epipolar loss is defined using the Sampson
residual~\cite{hartley2004multiple}
\begin{equation}
	\label{eq:tuning_epipolar}
	\begin{aligned}
		\mathcal L_{\mathrm{epi}}
		&=
		\left\langle
		\gamma_{ts}^{k}\rho_{\mathrm H}(r_{\mathrm S})
		\right\rangle,\\
		r_{\mathrm S}
		&=
		\frac{
			\tilde{\mathbf u}_t^\top
			\mathbf F_{k,ts}
			\tilde{\mathbf u}_s
		}{
			\sqrt{
				\|(\mathbf F_{k,ts}\tilde{\mathbf u}_s)_{1:2}\|_2^2
				+
				\|(\mathbf F_{k,ts}^{\top}\tilde{\mathbf u}_t)_{1:2}\|_2^2
			}
		}.
	\end{aligned}
\end{equation}
Here, $\tilde{\mathbf u}_t$ and $\tilde{\mathbf u}_s$ denote the corresponding homogeneous image coordinates, centered at the image center and normalized by the image diagonal. The notation $(\cdot)_{1:2}$ selects the first two vector components, and $\langle\cdot\rangle$ denotes averaging over matched image pairs and their point correspondences. The weight $\gamma_{ts}^{k}$ is computed from the depth-normalized baseline within chunk $k$.

VGGT-Long++ minimizes the combined objective
\begin{equation}
\label{eq:tuning_objective}
\mathcal L_{\mathrm{tune}}
=
\mathcal L_{\mathrm{cons}}+
\lambda_e\mathcal L_{\mathrm{epi}}
\end{equation}
by updating LoRA adapters~\cite{hu2022lora} in the QKV projections
of selected global attention blocks, while keeping all pretrained model weights frozen.
The two predictions share the same adapter parameters and
jointly contribute gradients to each update.


%% file: fig/scale_graph.tex
\begin{figure}[t]
  \centering
  {\preservepdfgroup
  \includegraphics[width=\linewidth]{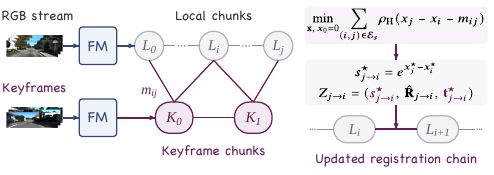}}
  \caption{\textbf{Scale-wise graph optimization.}
Keyframe chunks $K$ connect temporally separated local chunks $L$ through overlapping images. The optimized scales refine adjacent $\mathrm{Sim}(3)$ registrations, with translations recomputed accordingly. FM: foundation model.}
  \label{fig:scale_graph}
\end{figure}

%% file: sec/4_experiments.tex
\input{tables/trajectory}

\section{Experiments}
\label{sec:experiments}

\input{fig/trajectory_comparison}
\input{fig/dense_comparison}

\subsection{Experimental setup}
\label{sec:exp_setup}

\paragraph{Datasets and metrics.}
We evaluate trajectory accuracy on KITTI
Odometry~\cite{geiger2012kitti}, Oxford
Spires~\cite{tao2025spires,amap2026abot}, and VBR~\cite{brizi2024vbr}
using absolute trajectory error (ATE) RMSE in meters after a single
$\mathrm{Sim}(3)$ alignment of each complete trajectory to ground truth.
We further assess dense reconstruction qualitatively in outdoor and
indoor scenes and quantitatively on 7-Scenes~\cite{shotton2013scenecoordinate}
and NeuralRGBD~\cite{azinovic2022neuralrgbd}, reporting accuracy (Acc.)
and completeness (Comp.) in cm and F1 at a 5\,cm threshold.
All metrics are averaged across sequences.

\paragraph{Implementation.}
We use pretrained $\pi^3$~\cite{wang2025pi3} as the foundation model.
Local chunks contain 20 frames, with a 3-frame overlap between
consecutive chunks.
Keyframes are selected using $\tau_{\mathrm{kf}}=0.1\,\mathrm{rad}$
and grouped into 75-keyframe chunks with a 25-keyframe overlap.
Each loop chunk contains 40 frames from the two matched local chunks.
We insert rank-64 LoRA adapters into global attention blocks
10--15 and optimize them with AdamW~\cite{loshchilov2019adamw}
at a learning rate of $5\times10^{-4}$.
The loss weights are $\lambda_R=\lambda_t=\lambda_e=1$ and
$\lambda_K=0.5$.
For comparison, we evaluate the $\pi^3$-based variant of
VGGT-Long, denoted as VGGT-Long ($\pi^3$), with the same
input resolutions and local chunk settings as VGGT-Long++.
We use random seed 42 for all comparisons and ablation studies.
All experiments are conducted on a single NVIDIA RTX 4090 GPU
and an Intel Core Ultra 7 265K CPU.

\subsection{Camera trajectory accuracy}
\label{sec:exp_trajectory}

Tables~\ref{tab:trajectory} and~\ref{tab:trajectory_cross_benchmark}
show that VGGT-Long++ achieves the lowest mean ATE RMSE on
KITTI, Oxford Spires, and VBR.
Compared with VGGT-Long ($\pi^3$), it reduces mean errors by
68.3\%, 83.4\%, and 95.0\%, respectively, demonstrating the
effectiveness of the proposed framework with the same
foundation model.
VGGT-Long++ requires no additional offline training, yet
outperforms LingBot-Map and LoGeR$^\ast$, both trained for
long-sequence reconstruction.
It also achieves lower mean errors than SURE-Map + LC,
AMB3R-SLAM, and CLoSeR, which employ recent-segment scale
recalibration, a larger DA3 mapping backbone, and fast-weight
memory with loop closure, respectively.
Our scale graph uses new keyframe constraints to refine scales
throughout the accumulated registration chain, while
camera-guided online adaptation improves subsequent camera
predictions through geometric supervision.
Figure~\ref{fig:trajectory_comparison} further shows closer
agreement with ground truth over complete trajectories,
with reduced distortions on Bodleian-02 and Diag-0.
Evaluations with seeds 0--4 show consistent accuracy across
all three benchmarks, with detailed results provided in the
supplementary material.

\input{tables/trajectory_cross_benchmark}

\subsection{Dense reconstruction quality}
\label{sec:exp_geometry}

Figure~\ref{fig:dense_comparison} shows that VGGT-Long++
produces globally coherent reconstructions of KITTI 05 and
Bodleian-02.
Compared with VGGT-Long ($\pi^3$), the complete maps better
preserve the scene layout, while close-up views show reduced
structural duplication and clearer facade and window boundaries.
SURE-Map, CLoSeR, and LASER improve geometric quality through
uncertainty filtering, cross-window feature propagation, and
depth-layer alignment, respectively.
Our depth and affine alignment corrects both regional depth
and calibration discrepancies using overlapping images across
local, keyframe, and loop chunks.
Table~\ref{tab:geometry} further shows competitive reconstruction
quality on 7-Scenes and the best accuracy and F1 score on NeuralRGBD.

\input{tables/geometry}

\subsection{Ablation studies}
\label{sec:exp_components}

\input{fig/trajectory_ablation}

Table~\ref{tab:component_ablation} shows that scale-wise graph
optimization improves trajectory accuracy across all three benchmarks,
both with and without loop constraints.
Figure~\ref{fig:trajectory_ablation} illustrates the resulting
trajectory improvements.
All three keyframe-based formulations in
Table~\ref{tab:optimization_formulations} outperform the baseline
on KITTI, demonstrating the value of keyframe chunks as shared
references for long-range alignment.
Among them, scale-wise graph optimization achieves the lowest mean
ATE RMSE, supporting the benefit of optimizing scales separately
from poses.
Figure~\ref{fig:configuration_efficiency} shows that larger keyframe
chunks further improve accuracy at a fixed overlap, while the default
setting balances temporal coverage and scale update frequency.

\input{tables/component_ablation}
\input{tables/optimization_formulations}

\input{fig/geometry_ablation}
\input{fig/configuration_efficiency}

\begingroup
\widowpenalty=10000
With adapted predictions and global $\mathrm{Sim}(3)$ transformations
fixed, layer-wise depth adjustments reduce the mean depth-normalized 3D discrepancy between pointmaps of overlapping images on KITTI from 3.76\% to 2.67\%, and chunk-wise affine transformations further reduce it to 1.82\%. Figure~\ref{fig:geometry_ablation} illustrates reduced structural duplication along rooflines, picture-frame boundaries, and lampshade edges.
\par\endgroup

\Needspace{3\baselineskip}
Table~\ref{tab:component_ablation} shows that consistency supervision
improves trajectory accuracy on KITTI and Oxford Spires.
Epipolar supervision adds constraints from image correspondences
to address shared prediction errors, further improving accuracy
on KITTI and VBR.
With both losses, the adapted model outperforms its unadapted
counterpart across all three benchmarks while avoiding the
separate adaptation forward passes used by
VGGT-Align~\cite{zhang2026vggtalign}.
Figure~\ref{fig:trajectory_ablation} further shows closer agreement
with ground truth after online adaptation.
Additional ablation results are provided in the supplementary material.

\input{tables/runtime}

\Needspace{8\baselineskip}
\subsection{Efficiency analysis}
\label{sec:exp_runtime}

VGGT-Long++ achieves high reconstruction accuracy while maintaining
an average streaming throughput of 28.86 FPS across KITTI sequences,
with a peak GPU memory allocation of 13.85\,GiB.
Table~\ref{tab:runtime} shows the per-frame
times of the main pipeline stages.

%% file: tables/trajectory.tex
\begin{table*}[!tbp]
\centering
\captionsetup{skip=2.5pt}
\caption{\textbf{Trajectory accuracy on KITTI Odometry.}
ATE RMSE (m, $\downarrow$).
Calibrated: known intrinsics, LC: loop closure.
Best/second-best per group: \textbf{bold}/\underline{underline}.
$\dagger$: our reproductions.
TL: tracking lost; excluded from the mean.
VGGT-Long variants are labeled by foundation model:
$\pi^3$ or Depth Anything 3 (DA3).
TTT3R/LoGeR$^\ast$: state resets; $^\ast$: feedforward pose alignment.}
\label{tab:trajectory}
\fontsize{7}{7.6}\selectfont
\setlength{\tabcolsep}{6.5pt}
\renewcommand{\arraystretch}{0.96}
\setlength{\bigstrutjot}{2.2pt}
\setlength{\aboverulesep}{0.35ex}
\setlength{\belowrulesep}{0.5ex}
\begin{tabular}{cl*{12}{r}}
\toprule
 & Method & \textbf{00} & \textbf{01} & \textbf{02} & \textbf{03} & \textbf{04} & \textbf{05} & \textbf{06} & \textbf{07} & \textbf{08} & \textbf{09} & \textbf{10} & \textbf{Mean} \\
 & \textcolor{black!65}{Length (m)} & \textcolor{black!65}{3,720} & \textcolor{black!65}{2,450} & \textcolor{black!65}{5,070} & \textcolor{black!65}{560} & \textcolor{black!65}{390} & \textcolor{black!65}{2,210} & \textcolor{black!65}{1,230} & \textcolor{black!65}{690} & \textcolor{black!65}{3,220} & \textcolor{black!65}{1,710} & \textcolor{black!65}{920} & \\
\midrule
\multirow{3}[6]{*}{\rotatebox{90}{\shortstack{\textit{Calibrated}\\\textit{VO / SLAM}}}} & DPV-SLAM~\cite{lipson2024dpvslam}\bigstrut & 113.19 & \underline{16.76} & \underline{113.06} & \underline{2.44} & \underline{0.99} & \underline{59.42} & \underline{53.32} & \underline{18.90} & \underline{110.36} & 74.60 & \underline{13.74} & \underline{52.43} \\
 & DPVO~\cite{teed2023dpvo}\bigstrut & \underline{113.17} & \textbf{16.56} & 113.37 & 2.45 & \textbf{0.98} & 59.59 & 55.63 & 19.40 & 110.64 & \underline{74.44} & 13.80 & 52.73 \\
 & DROID-W~\cite{li2026droidw}\bigstrut & \textbf{48.26} & 207.57 & \textbf{52.80} & \textbf{1.76} & 4.65 & \textbf{25.16} & \textbf{5.34} & \textbf{3.35} & \textbf{29.05} & \textbf{35.56} & \textbf{8.01} & \textbf{38.32} \\
\midrule
\multirow{9}{*}{\rotatebox{90}{\shortstack{\textit{Uncalibrated}\\\textit{Offline}}}} & Scal3R~\cite{xie2026scal3r} & \underline{4.30} & \textbf{45.29} & 42.06 & 3.36 & 1.74 & 3.30 & \underline{2.49} & 2.03 & 36.69 & 12.32 & \textbf{6.46} & 14.55 \\
 & VGGT-Long~\cite{deng2026vggtlong} & 8.63 & 58.37 & 52.79 & 8.83 & 4.23 & 9.83 & 4.65 & 2.67 & 73.12 & 31.84 & 27.67 & 25.69 \\
 & VGGT-Long ($\pi^3$)$^{\dagger}$~\cite{deng2026vggtlong,wang2025pi3} & 12.95 & 111.96 & 82.96 & 10.12 & 1.36 & 9.16 & 5.75 & 3.44 & 68.65 & 26.33 & 33.45 & 33.29 \\
 & VGGT-Long (DA3)~\cite{deng2026vggtlong,lin2025depthanything3} & 5.13 & 78.76 & 35.64 & 5.38 & 3.18 & \underline{3.04} & 2.83 & 2.32 & \underline{26.55} & 8.86 & 13.42 & 16.83 \\
 & VGGT-Align~\cite{zhang2026vggtalign} & 7.74 & 65.11 & 47.04 & 5.12 & 2.55 & 4.97 & 4.21 & 4.48 & 43.98 & 23.11 & 11.66 & 19.99 \\
 & VGGT-Bridge~\cite{choi2026vggtbridge} & 13.74 & 56.34 & 48.70 & 9.33 & 4.00 & 11.02 & 6.93 & 3.15 & 47.15 & 26.24 & 22.05 & 22.61 \\
 & Anchor3R-Offline~\cite{tao2026anchor3r} & 19.68 & \underline{48.62} & 75.75 & 4.22 & 1.90 & 7.56 & 6.36 & 7.63 & 49.34 & 49.02 & 8.00 & 25.28 \\
 & GRF-Recon~\cite{li2026grfrecon} & 4.35 & 76.52 & \textbf{16.50} & \underline{1.65} & \underline{0.85} & 3.25 & \textbf{2.35} & \textbf{1.64} & \textbf{26.40} & \underline{7.25} & \underline{6.55} & \underline{13.39} \\
 & Glob3R~\cite{deng2026glob3r} & \textbf{2.77} & 53.12 & \underline{30.10} & \textbf{1.53} & \textbf{0.75} & \textbf{2.80} & 2.67 & \underline{1.77} & 32.76 & \textbf{5.51} & 11.51 & \textbf{13.21} \\
\midrule
\multirow{13}{*}{\rotatebox{90}{\shortstack{\textit{Uncalibrated}\\\textit{Streaming-based}}}} & VGGT-SLAM~\cite{maggio2025vggtslam} & 27.99 & 214.23 & 260.88 & 53.50 & 26.02 & 99.61 & 23.62 & 64.52 & 243.87 & 169.10 & 80.13 & 114.86 \\
 & VGGT-SLAM 2.0~\cite{maggio2026vggtslam2} & TL & 163.65 & TL & 50.04 & 19.38 & 159.58 & 46.35 & 57.80 & TL & 167.96 & 76.99 & 92.72 \\
 & TTT3R~\cite{chen2026ttt3r} & 118.19 & 102.98 & 242.05 & 14.37 & 3.52 & 31.62 & 37.66 & 11.99 & 81.52 & 79.10 & 29.81 & 68.44 \\
 & LingBot-Map~\cite{chen2026lingbot} & 33.57 & 151.48 & 66.02 & 2.72 & 2.09 & 12.79 & 6.06 & 4.82 & \textbf{13.36} & 19.50 & \underline{8.01} & 29.13 \\
 & HorizonStream + LC~\cite{cheng2026horizonstream} & 15.45 & \underline{24.15} & 71.01 & 5.53 & \underline{0.72} & 7.84 & 7.54 & 3.14 & 22.63 & 27.45 & 12.35 & 17.98 \\
 & LoGeR$^\ast$~\cite{zhang2026loger} & 30.47 & 47.91 & 36.32 & 5.38 & 1.95 & 26.34 & 6.60 & 5.55 & 24.41 & 10.12 & 10.11 & 18.65 \\
 & SLAMFormer-$\infty$~\cite{fang2026slamformer} & 15.39 & 96.58 & 28.81 & 4.97 & 4.16 & 11.36 & 13.54 & 8.53 & 37.00 & 18.85 & 13.93 & 23.01 \\
 & LASER~\cite{ding2026laser} & 6.14 & 121.61 & 59.87 & 2.64 & 1.36 & \underline{2.73} & \textbf{2.92} & 2.28 & 33.14 & 17.95 & 15.20 & 24.17 \\
 & ABot-Recon + LC~\cite{amap2026abot} & 11.70 & \textbf{21.55} & 39.00 & 4.38 & 2.93 & 6.00 & 8.70 & \textbf{1.65} & 22.30 & 13.99 & 16.17 & 13.49 \\
 & SURE-Map + LC~\cite{liu2026suremap} & 20.59 & 39.54 & 35.85 & \underline{2.28} & 1.33 & 17.48 & 3.36 & 3.99 & \underline{14.38} & 17.87 & 10.20 & 15.17 \\
 & AMB3R-SLAM~\cite{wang2026amb3rslam} & \textbf{3.67} & 55.91 & 31.55 & 4.80 & 2.04 & \textbf{2.68} & \underline{3.33} & \underline{1.73} & 23.56 & \underline{9.03} & \textbf{5.96} & 13.11 \\
 & CLoSeR~\cite{li2026closer} & 6.58 & 41.80 & \underline{28.38} & 4.88 & 1.82 & 6.58 & 4.77 & 4.87 & 26.51 & \textbf{5.64} & 8.83 & \underline{12.79} \\
\addlinespace[1pt]
\rowcolor[rgb]{0.965,0.950,0.977}
 & \textbf{VGGT-Long++ (ours)} & \underline{4.50} & 34.14 & \textbf{20.04} & \textbf{1.49} & \textbf{0.50} & 3.09 & 3.91 & 3.32 & 22.39 & 12.09 & 10.53 & \textbf{10.55} \\
\bottomrule
\end{tabular}
\end{table*}

%% file: fig/trajectory_comparison.tex
\begin{figure}[t]
\centering
\captionsetup{skip=2.5pt}
\includegraphics[width=0.96\linewidth]{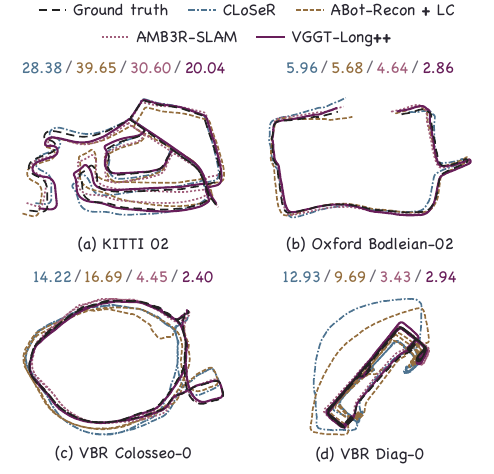}
\caption{\textbf{Camera trajectory comparison.}
Colored values report ATE RMSE (m).
Baseline trajectories are from our reproductions.}
\label{fig:trajectory_comparison}
\end{figure}

%% file: fig/dense_comparison.tex
\begin{figure*}[t]
\centering
\includegraphics[width=0.98\textwidth]{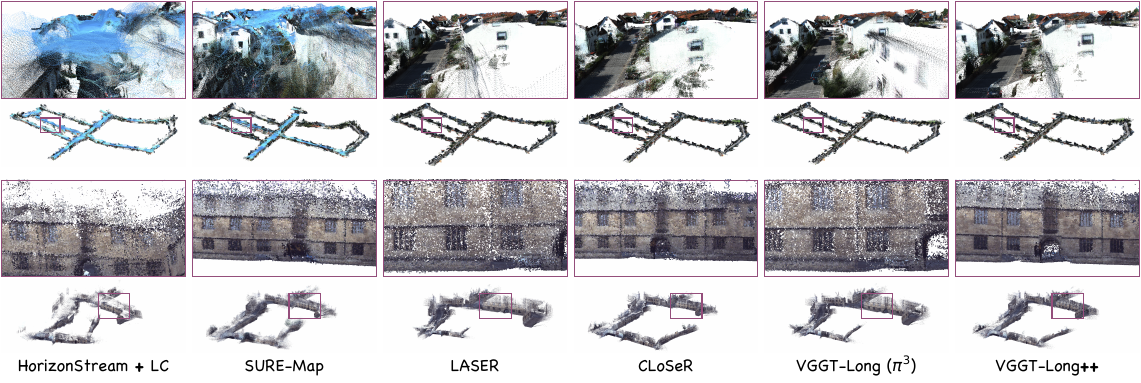}
\caption{\textbf{Dense reconstruction comparison.}
KITTI 05 (top two rows) and Oxford Spires Bodleian-02
(bottom two rows).
Close-up views of the boxed regions are shown above the full-sequence
reconstructions. All methods are rendered from matched viewpoints.}
\label{fig:dense_comparison}
\end{figure*}

%% file: tables/trajectory_cross_benchmark.tex
\begin{table}[!htbp]
\centering
\caption{\textbf{Trajectory accuracy on Oxford Spires and VBR.}
Mean~ATE~RMSE (m, $\downarrow$).
Best/second-best: \textbf{bold}/\underline{underline}.
\mbox{Our reproductions:} $\dagger$ Oxford; $\ddagger$ both benchmarks.}
\label{tab:trajectory_cross_benchmark}
\fontsize{7.5}{8.5}\selectfont
\setlength{\tabcolsep}{2pt}
\renewcommand{\arraystretch}{1.00}
\setlength{\aboverulesep}{0.3ex}
\setlength{\belowrulesep}{0.4ex}
\begin{tabular}{@{}lrr@{}}
\toprule
Method & \textbf{Oxford} & \textbf{VBR} \\
\midrule
DPVO~\cite{teed2023dpvo} & \underline{1.56} & 72.25 \\
DROID-W~\cite{li2026droidw} & 2.27 & 73.29 \\
LingBot-Map~\cite{chen2026lingbot} & 7.32 & 29.45 \\
HorizonStream + LC~\cite{cheng2026horizonstream} & 7.93 & 17.64 \\
ABot-Recon + LC~\cite{amap2026abot} & 4.02 & 9.99 \\
SURE-Map + LC~\cite{liu2026suremap} & 4.63 & 22.12 \\
AMB3R-SLAM$^{\dagger}$~\cite{wang2026amb3rslam} & 2.00 & 7.42 \\
AMB3R-SLAM ($\Omega$)$^{\dagger}$~\cite{wang2026amb3rslam} & 2.30 & \underline{7.15} \\
CLoSeR$^{\dagger}$~\cite{li2026closer} & 4.00 & 11.15 \\
VGGT-Long ($\pi^3$)$^{\ddagger}$~\cite{deng2026vggtlong,wang2025pi3} & 8.49 & 72.20 \\
\addlinespace[1pt]
\rowcolor[rgb]{0.965,0.950,0.977}
\textbf{VGGT-Long++ (ours)} & \textbf{1.41} & \textbf{3.59} \\
\bottomrule
\end{tabular}
\end{table}

%% file: tables/geometry.tex
\begin{table}[!htbp]
\centering
\caption{\textbf{7-Scenes and NeuralRGBD reconstruction quality.}
Acc./Comp. (cm, $\downarrow$),
F1 at 5\,cm (\%, $\uparrow$).
$\dagger$: our reproductions.}
\label{tab:geometry}
\fontsize{7.5}{8.5}\selectfont
\setlength{\tabcolsep}{1.8pt}
\renewcommand{\arraystretch}{1.00}
\setlength{\aboverulesep}{0.3ex}
\setlength{\belowrulesep}{0.4ex}
\begin{tabular}{@{}lrrrrrr@{}}
\toprule
& \multicolumn{3}{c}{\textbf{7-Scenes}} & \multicolumn{3}{c}{\textbf{NeuralRGBD}} \\
\cmidrule(lr){2-4}\cmidrule(l){5-7}
Method & Acc. & Comp. & F1 & Acc. & Comp. & F1 \\
\midrule
LingBot-Map~\cite{chen2026lingbot} & \underline{3.00} & 4.00 & 82.38 & 7.00 & 3.00 & 65.10 \\
SURE-Map~\cite{liu2026suremap} & 3.30 & 4.50 & 81.93 & 6.70 & 3.50 & 66.20 \\
CLoSeR$^{\dagger}$~\cite{li2026closer} & \textbf{2.81} & \textbf{2.78} & \textbf{87.81} & 7.68 & 2.39 & 68.14 \\
LASER$^{\dagger}$~\cite{ding2026laser} & 3.61 & 3.10 & 83.61 & 4.95 & \textbf{1.98} & 81.05 \\
VGGT-Long ($\pi^3$)$^{\dagger}$~\cite{deng2026vggtlong,wang2025pi3} & 3.56 & 3.20 & \underline{83.82} & \underline{4.74} & 2.51 & \underline{82.11} \\
\midrule
\rowcolor[rgb]{0.965,0.950,0.977}
\textbf{VGGT-Long++ (ours)} & 3.56 & \underline{2.98} & 83.57 & \textbf{4.61} & \underline{2.20} & \textbf{84.02} \\
\bottomrule
\end{tabular}
\end{table}

%% file: fig/trajectory_ablation.tex
\begin{figure*}[t]
\centering
\includegraphics[width=0.96\textwidth]{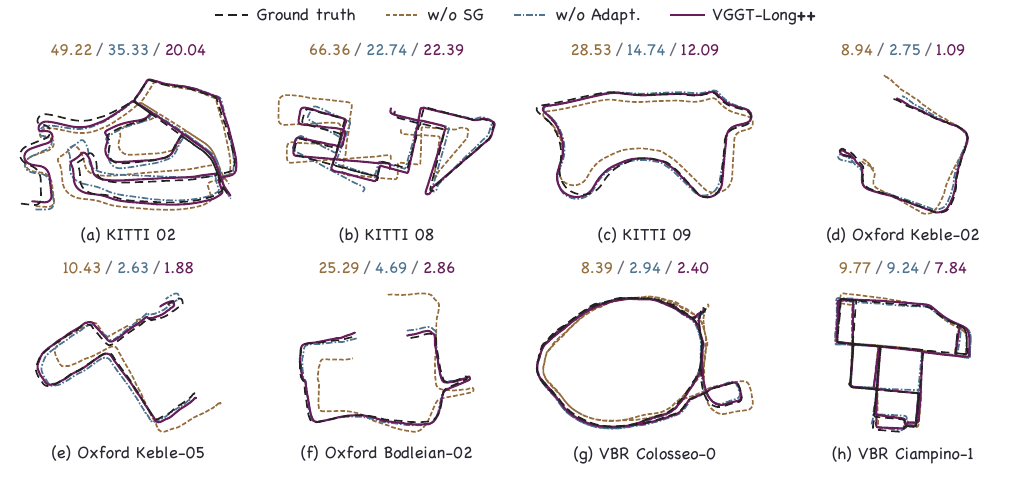}
\caption{\textbf{Trajectory accuracy ablations.}
SG: scale-wise graph optimization; Adapt.: camera-guided online adaptation.
Colored values report ATE RMSE (m).}
\label{fig:trajectory_ablation}
\end{figure*}

%% file: tables/component_ablation.tex
\begin{table}[!htbp]
\centering
\caption{\textbf{Trajectory accuracy ablations.}
Mean ATE RMSE (m, $\downarrow$).
SG: scale-wise graph optimization; LC: loop closure.
\mbox{Shaded row: full method.}}
\label{tab:component_ablation}
\fontsize{7.5}{8.5}\selectfont
\setlength{\tabcolsep}{1.8pt}
\renewcommand{\arraystretch}{1.00}
\setlength{\aboverulesep}{0.3ex}
\setlength{\belowrulesep}{0.4ex}
\renewcommand{\cmark}{\raisebox{0.3pt}{\ding{51}}}
\renewcommand{\xmark}{\raisebox{0.3pt}{\ding{55}}}
\begin{tabular}{@{}ccccrrr@{}}
\toprule
SG & LC & $\mathcal L_{\mathrm{cons}}$ & $\mathcal L_{\mathrm{epi}}$ & \textbf{KITTI} & \textbf{Oxford} & \textbf{VBR} \\
\midrule
\xmark & \xmark & \cmark & \cmark & 52.80 & 9.91 & 72.60 \\
\cmark & \xmark & \cmark & \cmark & 16.80 & 1.93 & 27.69 \\
\xmark & \cmark & \cmark & \cmark & 31.35 & 7.46 & 7.24 \\
\midrule
\cmark & \cmark & \xmark & \xmark & 15.40 & 2.05 & \underline{3.81} \\
\cmark & \cmark & \cmark & \xmark & \underline{10.71} & \textbf{1.37} & 4.01 \\
\rowcolor[rgb]{0.965,0.950,0.977}
\cmark & \cmark & \cmark & \cmark & \textbf{10.55} & \underline{1.41} & \textbf{3.59} \\
\bottomrule
\end{tabular}
\end{table}

%% file: tables/optimization_formulations.tex
\begin{table}[!htbp]
\centering
\caption{\textbf{Keyframe-based optimization ablations.}
	KITTI ATE RMSE (m).
	Predictions and loop constraints are fixed.
	--: no keyframe constraints;
	SG: scale-wise graph optimization.}
\label{tab:optimization_formulations}
\fontsize{7.5}{8.5}\selectfont
\setlength{\tabcolsep}{2pt}
\renewcommand{\arraystretch}{1.00}
\begin{tabular}{@{}lr@{}}
\toprule
Formulation & \textbf{Mean ATE} $\downarrow$ \\
\midrule
-- & 31.35 \\
\midrule
$\mathrm{Sim}(3)$ & 28.42 \\
Joint (scale + $\mathrm{Sim}(3)$) & \underline{11.21} \\
\rowcolor[rgb]{0.965,0.950,0.977}
SG (ours) & \textbf{10.55} \\
\bottomrule
\end{tabular}
\end{table}

%% file: fig/geometry_ablation.tex
\begin{figure}[htbp]
\centering
\includegraphics[width=\linewidth]{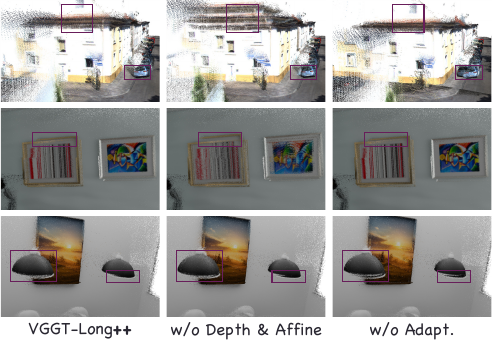}
\caption{\textbf{Reconstruction quality ablations.}
KITTI (top) and NeuralRGBD (middle and bottom).
Adapt.: camera-guided online adaptation.
Viewpoints and boxes match across columns.}
\label{fig:geometry_ablation}
\end{figure}

%% file: fig/configuration_efficiency.tex
\begin{figure}[!htbp]
\centering
\includegraphics[scale=0.98,trim=7bp 1bp 15bp 3bp,clip]{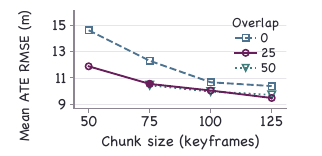}
\caption{\textbf{Keyframe chunk settings.}
Results are averaged over KITTI sequences. The filled circle marks the default setting.}
\label{fig:configuration_efficiency}
\end{figure}

%% file: tables/runtime.tex
\begin{table}[!htbp]
\centering
\caption{\textbf{Runtime breakdown.}
	Mean time on KITTI (ms/frame).
	Local, Keyframe, and Loop: inference;
	Adapt.: camera-guided online adaptation;
	SG: scale-wise graph optimization.}
\label{tab:runtime}
\fontsize{7.5}{8.5}\selectfont
\setlength{\tabcolsep}{2pt}
\renewcommand{\arraystretch}{1.00}
\setlength{\aboverulesep}{0.3ex}
\setlength{\belowrulesep}{0.4ex}
\begin{tabular}{@{}cccccc@{}}
\toprule
\textbf{Local + Adapt.} & \textbf{Keyframe} & \textbf{Loop} & \textbf{SG} & \textbf{Global} {\boldmath$\mathrm{Sim}(3)$} & \textbf{Depth \& Affine} \\
\midrule
19.88 & 2.48 & 0.75 & 1.77 & 0.31 & 10.48 \\
\bottomrule
\end{tabular}
\end{table}

%% file: sec/5_conclusion.tex
\section{Conclusion}
\label{sec:conclusion}

In this paper, we presented VGGT-Long++, extending VGGT-Long for globally
consistent long-sequence 3D reconstruction.
By refining the registration chain of local chunks, correcting residual geometric
discrepancies, and improving subsequent model predictions,
VGGT-Long++ reduces the propagation of prediction errors from
local chunks into the global reconstruction.
Experiments demonstrate state-of-the-art trajectory accuracy,
improved dense reconstruction quality, and efficient streaming
execution without additional offline training.
Future work will explore extending the framework to more capable
foundation models.

%% file: supplementary.tex
\clearpage
\appendix
\setcounter{equation}{0}
\setcounter{figure}{0}
\setcounter{table}{0}
\setcounter{algorithm}{0}
\renewcommand{\theequation}{S\arabic{equation}}
\renewcommand{\thefigure}{S\arabic{figure}}
\renewcommand{\thetable}{S\arabic{table}}
\renewcommand{\thealgorithm}{S\arabic{algorithm}}

\renewcommand{\theHsection}{supp.\Alph{section}}
\renewcommand{\theHsubsection}{\theHsection.\arabic{subsection}}
\renewcommand{\theHequation}{supp.\arabic{equation}}
\renewcommand{\theHfigure}{supp.\arabic{figure}}
\renewcommand{\theHtable}{supp.\arabic{table}}
\providecommand{\theHalgorithm}{}
\renewcommand{\theHalgorithm}{supp.\arabic{algorithm}}

\setlength{\dblfloatsep}{8pt plus 2pt minus 2pt}
\setlength{\dbltextfloatsep}{8pt plus 2pt minus 2pt}
\setlength{\floatsep}{12pt plus 2pt minus 2pt}
\setlength{\textfloatsep}{8pt plus 2pt minus 2pt}
\setlength{\intextsep}{8pt plus 2pt minus 2pt}
\renewcommand{\textfraction}{0.15}
\renewcommand{\dbltopfraction}{0.85}
\renewcommand{\dblfloatpagefraction}{0.75}
\setcounter{dbltopnumber}{3}
\makeatletter
\setlength{\@dblfptop}{0pt}
\setlength{\@dblfpsep}{12pt}
\setlength{\@dblfpbot}{0pt plus 1fil}
\makeatother

\maketitlesupplementary
\pdfbookmark[0]{Supplementary Material}{supplementary}
\etocdepthtag.toc{supplement}
\begingroup
\fontsize{9.4}{12}\selectfont
\hypersetup{linktoc=all,linkcolor=black}
\etocsettagdepth{main}{none}
\etocsettagdepth{supplement}{subsection}
\etocsettocstyle{\section*{Contents}}{}
\tableofcontents
\endgroup
\input{sec/X_suppl}

%% file: sec/X_suppl.tex
\input{sec/supplement/method_details}
\input{sec/supplement/experiments}

\newpage
\input{sec/supplement/other_methods}
\input{sec/supplement/reconstructions}

%% file: sec/supplement/method_details.tex
\section{Additional method details}
\label{sec:supp_method}

\subsection{Chunk-wise \texorpdfstring{$\mathrm{Sim}(3)$}{Sim(3)} registration}
\label{sec:supp_chunk_registration}

VGGT-Long++ estimates the relative $\mathrm{Sim}(3)$
transformation from chunk $j$ to chunk $i$ by first estimating
scale from depth predictions for overlapping images and then
aligning the corresponding camera poses.
For each $I_t\in\mathcal O_{ij}$, the two confidence maps define a shared
threshold
\begin{equation}
\tau_{ij,t}
=0.1\min_{k\in\{i,j\}}
\left(\operatorname{med}_{v}\chi_{k,t}(v)\right).
\end{equation}
The mask $\Omega_{ij,t}$ retains pixels $u$ with positive depth in both
predictions and $\min\{\chi_{i,t}(u),\chi_{j,t}(u)\}>\tau_{ij,t}$.
The relative scale is then $\hat s_{j\rightarrow i}=\exp(m_{ij})$, where
$m_{ij}$ is computed using Eq.~\mainequation{eq:scale_measurement} of the main paper.

For each overlapping image, the camera-to-chunk rotations yield a relative
rotation $\mathbf R_{i,t}\mathbf R_{j,t}^{\top}$ from chunk $j$ to chunk $i$.
VGGT-Long++ projects their sum onto $\mathrm{SO}(3)$ using singular value
decomposition (SVD) to obtain
\begin{equation}
\hat{\mathbf R}_{j\rightarrow i}
=
\operatorname{proj}_{\mathrm{SO}(3)}
\left(
\sum_{I_t\in\mathcal O_{ij}}
\mathbf R_{i,t}\mathbf R_{j,t}^{\top}
\right).
\end{equation}
With $\hat s_{j\rightarrow i}$ and $\hat{\mathbf R}_{j\rightarrow i}$ fixed,
translation is estimated from the corresponding camera centers as in
Eq.~\mainequation{eq:registration_update} of the main paper, using
$\hat s_{j\rightarrow i}$ in place of $s^\star_{j\rightarrow i}$.
The same registration procedure is used between adjacent local chunks and
between each loop chunk and its two matched local chunks.

\subsection{Loop detection and verification}
\label{sec:supp_loop_detection}

For each incoming image $I_t$, VGGT-Long++ uses SALAD to retrieve
the five most similar images $I_s$ satisfying
$t-s>\tau_{\mathrm{time}}$, where $\tau_{\mathrm{time}}=80$
corresponds to four times the local chunk size.
Each retrieved match $(I_t,I_s)$ with similarity above
$\tau_{\mathrm{sim}}=0.75$ defines a loop candidate between
the two existing local chunks containing these images.
For each nonadjacent chunk pair, only the highest-scoring match
is retained as its representative for subsequent filtering.

Revisits after a trajectory detour should have a small endpoint
distance relative to the intervening path length.
A candidate between local chunks $i$ and $j$, with $i<j$,
therefore passes the chord-to-arc ratio filter if
\begin{equation}
\label{eq:supp_chord_arc}
\frac{\|\mathbf a_j-\mathbf a_i\|_2}
{\displaystyle\sum_{k=i}^{j-1}
\|\mathbf a_{k+1}-\mathbf a_k\|_2}
<\tau_a,
\qquad \tau_a=0.85.
\end{equation}
Here, $\mathbf a_k$ denotes the center of the first camera
in local chunk $k$, expressed in the world frame.

A valid loop should also yield consistent relative rotations
from the aligned camera poses and the matched images.
For the representative images $I_s\in\mathcal C_i$ and
$I_t\in\mathcal C_j$, the relative rotation from camera $s$
to camera $t$ is
\begin{equation}
\label{eq:supp_trajectory_rotation}
\mathbf Q^{\mathrm{traj}}
=
(\mathbf R^{\mathrm w}_{j,t})^\top
\mathbf R^{\mathrm w}_{i,s}.
\end{equation}
Here, $\mathbf R^{\mathrm w}_{i,s}$ and
$\mathbf R^{\mathrm w}_{j,t}$ denote their camera-to-world rotations.
The corresponding image-based rotation $\mathbf Q^{\mathrm{img}}$
is then recovered from SIFT matches and current intrinsics
through essential-matrix estimation with RANSAC.
The candidate passes the rotation consistency filter if
\begin{equation}
\label{eq:supp_rotation_consistency}
d_{\mathrm{SO}(3)}
\left(\mathbf Q^{\mathrm{traj}},\mathbf Q^{\mathrm{img}}\right)
\leq\tau_R,
\qquad \tau_R=90^\circ.
\end{equation}

Finally, candidates passing both filters undergo causal suppression,
with higher-similarity candidates considered first within each
new local chunk.
A candidate $(i,j)$ is suppressed if a previously accepted loop
$(i',j')$ satisfies
\begin{equation}
\label{eq:supp_causal_suppression}
\max\left(|i-i'|,\;|j-j'|\right)\leq r_{\mathrm{nms}},
\qquad r_{\mathrm{nms}}=2.
\end{equation}
For each accepted loop, all images in the two corresponding
local chunks are jointly reconstructed as a loop chunk,
which is registered using the procedure in
Sec.~\ref{sec:supp_chunk_registration}.

\subsection{Keyframe chunks at reduced resolution}
\label{sec:supp_keyframe_inference}

VGGT-Long++ reconstructs keyframe chunks at reduced resolution to lower
the computational and memory costs of joint inference.
Images are resized to a width of 336 pixels, with the height scaled
proportionally and rounded to the nearest multiple of 14.

For overlapping images, VGGT-Long++ constructs scale and
depth constraints on the keyframe prediction grid.
Depth and confidence maps from local chunks are therefore
resampled to this grid using bilinear interpolation, while
depth-layer labels are resampled using nearest-neighbor
interpolation.

\subsection{More details on depth adjustments}
\label{sec:supp_depth_adjustments}

For depth layers $a$ and $b$ predicted for the same image across
chunks, the median discrepancy $\eta_{ab}^{d}$ constrains
their relative adjustment $\delta_b-\delta_a$, with weight
\begin{equation}
\label{eq:supp_overlap_depth_weight}
w_{ab}=\frac{|\Omega_{ab}|}{HW}.
\end{equation}
Here, $\Omega_{ab}=M_a\cap M_b$ is the intersection of the
corresponding layer masks, and $H$ and $W$ denote the height
and width of the common pixel grid.

Temporal constraints encourage consistent depth adjustments
between consecutive frames within a chunk.
For layers $a$ and $b$ in these frames with
$\operatorname{IoU}(M_a,M_b)\geq0.3$, $\eta_{ab}^{d}$ is set
to zero and the weight is
\begin{equation}
\label{eq:supp_temporal_depth_weight}
w_{ab}
=
\frac{|\Omega_{ab}|}{HW}
\operatorname{IoU}(M_a,M_b).
\end{equation}

\subsection{More details on online adaptation}
\label{sec:supp_online_adaptation}

\paragraph{Loss weighting.}
VGGT-Long++ downweights frame pairs with small depth-normalized
baselines to reduce the influence of unstable translation directions.
For adjacent local chunks $i$ and $j$, let $z_k^{\mathrm{cons}}$ and
$z_k^{\mathrm{epi}}$ denote the median depths over $\mathcal O_{ij}$
and $\mathcal C_k$, respectively, for $k\in\{i,j\}$.
Each is computed as the median of per-image median depths.
Define
\begin{equation}
\label{eq:supp_baseline_weight}
g_{ts}^{k}(z)
=
\frac{\|\mathbf c_{k,s}-\mathbf c_{k,t}\|_2^2}
{\|\mathbf c_{k,s}-\mathbf c_{k,t}\|_2^2+(\tau z)^2},
\qquad \tau=0.01.
\end{equation}
The weights $\gamma_{ts}$ in $\mathcal L_t$ and $\gamma_{ts}^{k}$ in
$\mathcal L_{\mathrm{epi}}$ are
\begin{equation}
\label{eq:supp_loss_weights}
\begin{aligned}
\gamma_{ts}
&=
g_{ts}^{i}(z_i^{\mathrm{cons}})
g_{ts}^{j}(z_j^{\mathrm{cons}}),\\
\gamma_{ts}^{k}
&=
g_{ts}^{k}(z_k^{\mathrm{epi}}).
\end{aligned}
\end{equation}
The median depths and both weights are held fixed during backpropagation.

For input images of size $H\times W$, the Huber thresholds are $0.1$
for intrinsic consistency and $1/\sqrt{H^2+W^2}$ for epipolar
supervision.
Each Huber penalty is divided by its threshold.

\paragraph{Online updates.}
Algorithm~\ref{alg:supp_online_adaptation} summarizes the update procedure.
For epipolar supervision, SIFT matches are computed within
each local chunk among overlapping images and between these
images and their $\pm5$-frame neighbors.
The matches are then verified by RANSAC.
AdamW uses moment coefficients $(0.5,0.999)$ and weight decay $0.01$.

\begin{algorithm}[H]
\caption{Camera-guided online adaptation}
\label{alg:supp_online_adaptation}
\small
\begin{algorithmic}[1]
\Require Local chunks $\{\mathcal C_k\}_{k=0}^{M-1}$, foundation model
$f_\theta$ with trainable LoRA parameters $\theta$.
\Ensure Chunk predictions $\{\mathcal G_k\}$, adapted parameters $\theta$.
\State Initialize $\theta$ and AdamW.
\For{pairs of full local chunks $(i,j)=(0,1),(2,3),\ldots$}
\State Obtain image correspondences within $\mathcal C_i$
and $\mathcal C_j$.
\State $\mathcal G_i,\mathcal G_j\gets f_\theta(\mathcal C_i),f_\theta(\mathcal C_j)$.
\State Evaluate $\mathcal L_{\mathrm{cons}}$ and $\mathcal L_{\mathrm{epi}}$.
\State $\mathcal L_{\mathrm{tune}}\gets\mathcal L_{\mathrm{cons}}+\lambda_e\mathcal L_{\mathrm{epi}}$.
\State $\theta\gets\operatorname{AdamW}(\theta,\nabla_\theta\mathcal L_{\mathrm{tune}})$.
\State Output $\mathcal G_i,\mathcal G_j$.
\EndFor
\end{algorithmic}
\end{algorithm}

%% file: sec/supplement/experiments.tex
\begin{samepage}
\section{Additional experiments}
\label{sec:supp_visuals}

\subsection{More details on the experimental setup}
\label{sec:supp_exp_settings}

We use default input widths of 504 pixels for KITTI and VBR and
518 pixels for Oxford Spires, 7-Scenes, and NeuralRGBD.
Image heights are scaled proportionally and rounded to the nearest
multiple of 14.
We follow the trajectory and indoor dense reconstruction evaluation protocols
of LingBot-Map~\cite{chen2026lingbot}.\par
\end{samepage}

\subsection{Additional camera trajectory results}
\label{sec:supp_trajectory_accuracy}

\begin{figure*}[p]
\centering
\includegraphics[width=0.95\linewidth]{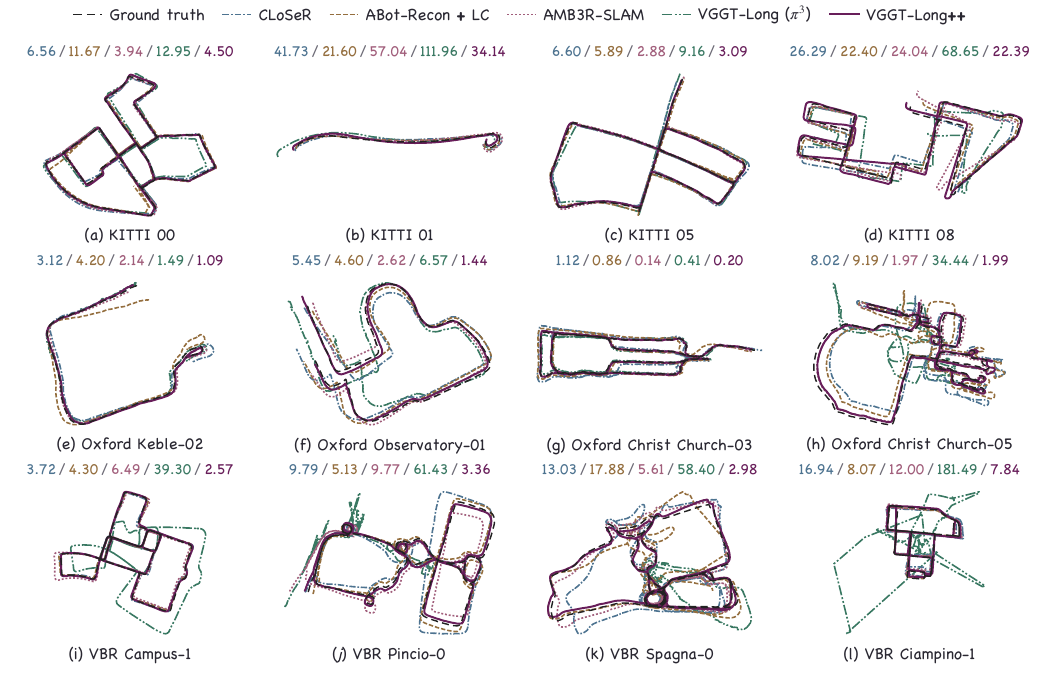}
\caption{\textbf{Additional camera trajectory comparisons.}
Colored values report ATE RMSE (m).
Baseline trajectories are from our reproductions.}
\label{fig:supp_trajectory_comparison}
\end{figure*}

Tables~\ref{tab:full_oxford_spires} and~\ref{tab:full_vbr}
show that VGGT-Long++ achieves the lowest mean ATE RMSE
among the compared methods on Oxford Spires and VBR,
with substantial improvements over VGGT-Long ($\pi^3$).
For SURE-Map + LC, we report the published means because
per-sequence results and the loop-closure implementation
are not publicly available.
Figure~\ref{fig:supp_trajectory_comparison} further illustrates
reduced large-scale trajectory distortions, particularly
on Pincio-0 and Spagna-0.

\input{tables/trajectory_oxford_spires_full}
\input{tables/trajectory_vbr_full}

\subsection{Trajectory accuracy across random seeds}
\label{sec:supp_seed_sensitivity}

Table~\ref{tab:supp_trajectory_seeds} shows that VGGT-Long++
consistently achieves high trajectory accuracy across random
seeds 0--4 on all three benchmarks, with standard deviations
ranging from 1.8\% to 5.5\% of the corresponding mean ATE RMSEs.

\input{tables/supplement/trajectory_seeds}

\begingroup
\makeatletter
\setlength{\@dblfpsep}{6pt}
\makeatother
\clearpage
\endgroup

\twocolumn[{
\begin{minipage}{\textwidth}
\centering
\captionsetup{hypcap=false}
\includegraphics[width=\linewidth]{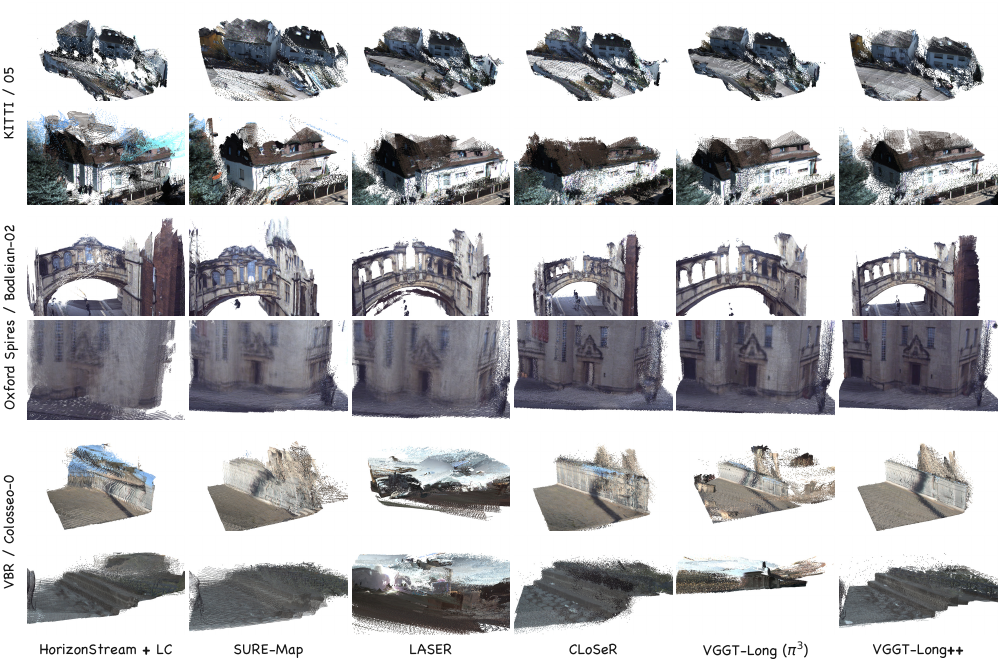}
\captionof{figure}{\textbf{Additional dense reconstruction comparison.}}
\label{fig:supp_dense_closeups}
\end{minipage}
\vspace{6pt}
}]

\subsection{Additional dense reconstruction results}
\label{sec:supp_dense_reconstructions}

We further evaluate long-sequence dense reconstruction on
Oxford Spires against laser-scanned surfaces visible from
the input cameras.
For each sequence, we align the reconstructed map to these
surfaces using a single global $\mathrm{Sim}(3)$ transformation
estimated from 3D point correspondences, followed by rigid
ICP refinement.
Table~\ref{tab:supp_oxford_dense_reconstruction} shows that
VGGT-Long++ achieves the lowest mean Chamfer distance and
the highest mean F1 score among the compared methods,
indicating closer agreement with measured scene geometry.

\input{tables/supplement/oxford_dense_reconstruction}

Figure~\ref{fig:supp_dense_closeups} illustrates improved
reconstruction quality across the three long sequences.
Compared with SURE-Map and CLoSeR, VGGT-Long++ reconstructs
clearer rooflines and window boundaries.
The improvement is particularly evident on Colosseo-0,
where VGGT-Long++ preserves distinct fence bars and terrace
steps while VGGT-Long ($\pi^3$) and LASER exhibit severe
geometric distortions.
Complete long-sequence reconstructions in
Figure~\ref{fig:supp_large_scale_reconstructions} complement
these local comparisons, with additional indoor results
presented in Figures~\ref{fig:supp_7scenes_maps}
and~\ref{fig:supp_nrgbd_maps}.

\subsection{Additional ablation studies}
\label{sec:supp_ablations}

\paragraph{Keyframe-based optimization.}
For the KITTI comparison in Table~\mainref{tab:optimization_formulations}
of the main paper, we hold chunk predictions and loop measurements fixed.
We compare scale-wise graph optimization with two alternative
formulations of keyframe constraints and a baseline using only
adjacent and loop constraints.
The $\mathrm{Sim}(3)$ graph formulation optimizes a similarity
transformation for each local and keyframe chunk, then uses
the optimized local transformations to update adjacent
registrations for subsequent global $\mathrm{Sim}(3)$
optimization with loop constraints.
The joint formulation instead optimizes the global
$\mathrm{Sim}(3)$ transformations of local chunks and the
log-scales of keyframe chunks, combining adjacent and loop
registration constraints with scale constraints from
keyframe chunks in a single objective.

\paragraph{Additional reconstruction quality ablations.}
We further assess geometric consistency by comparing the pointmaps
predicted for overlapping images in adjacent local chunks.
For chunks $i$ and $j$, the depth-normalized 3D discrepancy is
\begin{equation}
\label{eq:supp_pointmap_discrepancy}
D_{ij}
=
\underset{I_t\in\mathcal O_{ij}}{\operatorname{med}}\,
\underset{u\in\Omega^{\mathrm{eval}}_{ij,t}}{\operatorname{med}}
\frac{
  \left\|\mathbf P^{\mathrm{map}}_{i,t}(u)
        -\mathbf P^{\mathrm{map}}_{j,t}(u)\right\|_2
}{
  \tfrac12\bigl(\bar d_{i,t}(u)+\bar d_{j,t}(u)\bigr)
}.
\end{equation}
Here, $\mathbf P^{\mathrm{map}}$ denotes the pointmap in the world
frame for the evaluated variant, and $\bar d$ is the predicted
depth at the common scale before layer-wise adjustment.
All variants share the evaluation mask $\Omega^{\mathrm{eval}}_{ij,t}$,
which retains only pixels that pass confidence filtering in each chunk
both with and without online adaptation.
We average $D_{ij}$ over adjacent chunk pairs within each sequence
and then across sequences.

Table~\ref{tab:supp_geometry_consistency_ablation} shows that
camera-guided online adaptation reduces the mean discrepancy on KITTI.
With the adapted predictions and global $\mathrm{Sim}(3)$
transformations fixed, adding layer-wise depth adjustments
and then chunk-wise affine transformations yields further reductions.
These two steps also improve dense reconstruction quality on
NeuralRGBD, with gains in mean accuracy and F1 reported in
Table~\ref{tab:supp_indoor_geometry_ablation}.

\input{tables/supplement/geometry_consistency_ablation}
\input{tables/supplement/indoor_geometry_ablation}

%% file: tables/trajectory_oxford_spires_full.tex
\begin{table*}[p]
\centering
\caption{\textbf{Trajectory accuracy on Oxford Spires.}
ATE RMSE (m, $\downarrow$). Calibrated: known intrinsics, LC: loop closure.
Best/second-best per group: \textbf{bold}/\underline{underline}.
$\dagger$: our reproductions.
Dashes indicate unreported results.}
\label{tab:full_oxford_spires}
\begin{minipage}{0.87\linewidth}
\centering
\fontsize{7}{7.6}\selectfont
\setlength{\tabcolsep}{6.5pt}
\renewcommand{\arraystretch}{0.93}
\setlength{\bigstrutjot}{4.25pt}
\setlength{\aboverulesep}{0.35ex}
\setlength{\belowrulesep}{0.5ex}
\begin{tabularx}{\linewidth}{cX*{11}{r}}
\toprule
 & \multirow{2}{*}{Method} & \multicolumn{4}{c}{\textbf{Keble College}} & \multicolumn{2}{c}{\textbf{Observatory}} & \multicolumn{3}{c}{\textbf{Christ Church}} & \multicolumn{1}{c}{\textbf{Bodleian}} & \multirow{2}{*}{\textbf{Mean}} \\
\cmidrule(lr){3-6}\cmidrule(lr){7-8}\cmidrule(lr){9-11}\cmidrule(lr){12-12}
 & & \textbf{02} & \textbf{03} & \textbf{04} & \textbf{05} & \textbf{01} & \textbf{02} & \textbf{02} & \textbf{03} & \textbf{05} & \textbf{02} & \\
\midrule
\multirow{2}[tb4]{*}{\rotatebox[origin=c]{90}{\textit{Calibrated}}} & DPVO~\cite{teed2023dpvo}\bigstrut & \textbf{0.12} & \textbf{0.84} & \textbf{0.05} & \textbf{0.29} & \textbf{0.16} & \textbf{0.51} & \textbf{0.15} & \textbf{0.35} & \underline{12.04} & \textbf{1.08} & \textbf{1.56} \\
 & DROID-W~\cite{li2026droidw}\bigstrut & \underline{2.45} & \underline{0.99} & \underline{2.64} & \underline{4.27} & \underline{1.07} & \underline{1.06} & \underline{1.63} & \underline{0.75} & \textbf{1.56} & \underline{6.28} & \underline{2.27} \\
\midrule
\multirow{9}{*}{\rotatebox[origin=c]{90}{\textit{Uncalibrated}}} & LingBot-Map~\cite{chen2026lingbot} & 2.71 & 3.04 & 2.04 & 3.23 & 5.38 & 4.43 & 2.21 & 0.74 & 37.99 & 11.43 & 7.32 \\
 & HorizonStream + LC~\cite{cheng2026horizonstream} & 5.79 & 6.47 & 1.25 & 6.24 & 6.33 & 5.30 & 10.69 & 0.47 & 25.35 & 11.41 & 7.93 \\
 & ABot-Recon + LC~\cite{amap2026abot} & 4.24 & 1.70 & 2.04 & 4.44 & 4.60 & 4.40 & 3.07 & 0.85 & 9.03 & 5.76 & 4.02 \\
 & SURE-Map + LC~\cite{liu2026suremap} & \multicolumn{1}{c}{--} & \multicolumn{1}{c}{--} & \multicolumn{1}{c}{--} & \multicolumn{1}{c}{--} & \multicolumn{1}{c}{--} & \multicolumn{1}{c}{--} & \multicolumn{1}{c}{--} & \multicolumn{1}{c}{--} & \multicolumn{1}{c}{--} & \multicolumn{1}{c}{--} & 4.63 \\
 & AMB3R-SLAM$^{\dagger}$~\cite{wang2026amb3rslam} & 2.14 & \underline{0.69} & 0.89 & 2.40 & \underline{2.62} & \underline{3.11} & \underline{1.37} & \textbf{0.14} & \underline{1.97} & \underline{4.64} & \underline{2.00} \\
 & AMB3R-SLAM ($\Omega$)$^{\dagger}$~\cite{wang2026amb3rslam} & 2.07 & \textbf{0.50} & \textbf{0.24} & \textbf{1.81} & 4.91 & 4.45 & \textbf{1.09} & \underline{0.19} & \textbf{1.27} & 6.45 & 2.30 \\
 & CLoSeR$^{\dagger}$~\cite{li2026closer} & 3.12 & 2.92 & 1.40 & 3.89 & 5.45 & 5.79 & 2.30 & 1.12 & 8.02 & 5.95 & 4.00 \\
 & VGGT-Long ($\pi^3$)$^{\dagger}$~\cite{deng2026vggtlong,wang2025pi3} & \underline{1.49} & 2.84 & 1.77 & 5.27 & 6.57 & 6.33 & 3.49 & 0.41 & 34.44 & 22.28 & 8.49 \\
\addlinespace[1pt]
\rowcolor[rgb]{0.965,0.950,0.977}
 & \textbf{VGGT-Long++ (ours)} & \textbf{1.09} & 0.73 & \underline{0.64} & \underline{1.88} & \textbf{1.44} & \textbf{1.81} & 1.45 & 0.20 & 1.99 & \textbf{2.86} & \textbf{1.41} \\
\bottomrule
\end{tabularx}
\end{minipage}
\end{table*}

%% file: tables/trajectory_vbr_full.tex
\begin{table*}[p]
\centering
\caption{\textbf{Trajectory accuracy on VBR.}
ATE RMSE (m, $\downarrow$). Calibrated: known intrinsics, LC: loop closure.
Best/second-best per group: \textbf{bold}/\underline{underline}.
$\dagger$: our reproductions.
Dashes indicate unreported results.}
\label{tab:full_vbr}
\begin{minipage}{0.87\linewidth}
\centering
\fontsize{7}{7.6}\selectfont
\setlength{\tabcolsep}{6.5pt}
\renewcommand{\arraystretch}{0.93}
\setlength{\bigstrutjot}{4.25pt}
\setlength{\aboverulesep}{0.35ex}
\setlength{\belowrulesep}{0.5ex}
\begin{tabularx}{\linewidth}{cX*{8}{r}}
\toprule
 & Method & \textbf{Colosseo-0} & \textbf{Campus-0} & \textbf{Campus-1} & \textbf{Pincio-0} & \textbf{Spagna-0} & \textbf{Diag-0} & \textbf{Ciampino-1} & \textbf{Mean} \\
\midrule
\multirow{2}[tb4]{*}{\rotatebox[origin=c]{90}{\textit{Calibrated}}} & DPVO~\cite{teed2023dpvo}\bigstrut & \underline{90.42} & \textbf{109.72} & \textbf{58.49} & \underline{52.83} & \underline{57.84} & \underline{33.84} & \textbf{102.60} & \textbf{72.25} \\
 & DROID-W~\cite{li2026droidw}\bigstrut & \textbf{50.27} & \underline{125.02} & \underline{96.89} & \textbf{21.96} & \textbf{22.82} & \textbf{24.59} & \underline{171.52} & \underline{73.29} \\
\midrule
\multirow{9}{*}{\rotatebox[origin=c]{90}{\textit{Uncalibrated}}} & LingBot-Map~\cite{chen2026lingbot} & 11.11 & 21.45 & 12.39 & 46.30 & 35.40 & 16.16 & 63.36 & 29.45 \\
 & HorizonStream + LC~\cite{cheng2026horizonstream} & 24.66 & 27.22 & 8.72 & 16.29 & 21.33 & 7.36 & 17.92 & 17.64 \\
 & ABot-Recon + LC~\cite{amap2026abot} & 16.70 & 8.10 & 4.30 & 5.16 & 17.96 & 9.71 & 8.01 & 9.99 \\
 & SURE-Map + LC~\cite{liu2026suremap} & \multicolumn{1}{c}{--} & \multicolumn{1}{c}{--} & \multicolumn{1}{c}{--} & \multicolumn{1}{c}{--} & \multicolumn{1}{c}{--} & \multicolumn{1}{c}{--} & \multicolumn{1}{c}{--} & 22.12 \\
 & AMB3R-SLAM~\cite{wang2026amb3rslam} & 3.95 & 9.29 & 6.73 & 10.75 & 5.94 & 3.36 & 11.92 & 7.42 \\
 & AMB3R-SLAM ($\Omega$)~\cite{wang2026amb3rslam} & \underline{2.52} & 26.43 & 3.72 & \underline{3.95} & \underline{5.10} & \underline{2.66} & \textbf{5.66} & \underline{7.15} \\
 & CLoSeR~\cite{li2026closer} & 14.23 & \underline{7.43} & \underline{3.71} & 9.78 & 13.04 & 12.92 & 16.93 & 11.15 \\
 & VGGT-Long ($\pi^3$)$^{\dagger}$~\cite{deng2026vggtlong,wang2025pi3} & 68.87 & 93.26 & 39.30 & 61.43 & 58.40 & \textbf{2.65} & 181.49 & 72.20 \\
\addlinespace[1pt]
\rowcolor[rgb]{0.965,0.950,0.977}
 & \textbf{VGGT-Long++ (ours)} & \textbf{2.40} & \textbf{3.07} & \textbf{2.57} & \textbf{3.36} & \textbf{2.98} & 2.94 & \underline{7.84} & \textbf{3.59} \\
\bottomrule
\end{tabularx}
\end{minipage}
\end{table*}

%% file: tables/supplement/trajectory_seeds.tex
\begin{table}[H]
\centering
\caption{\textbf{Trajectory accuracy across random seeds.}
Mean ATE RMSE (m, $\downarrow$). SD denotes the sample standard deviation.}
\label{tab:supp_trajectory_seeds}
\fontsize{8}{9.5}\selectfont
\setlength{\tabcolsep}{5pt}
\renewcommand{\arraystretch}{1.05}
\begin{tabular}{lrrr}
\toprule
\textbf{Seed} & \textbf{KITTI} & \textbf{Oxford Spires} & \textbf{VBR}\\
\midrule
0 & 10.81 & 1.35 & 3.96\\
1 & 10.36 & 1.40 & 3.90\\
2 & 10.53 & 1.38 & 3.84\\
3 & 10.39 & 1.42 & 3.66\\
4 & 10.67 & 1.55 & 4.05\\
\midrule
\rowcolor[rgb]{0.965,0.950,0.977}
Mean $\pm$ SD & $10.55\pm0.19$ & $1.42\pm0.08$ & $3.88\pm0.15$\\
\bottomrule
\end{tabular}
\end{table}

%% file: tables/supplement/oxford_dense_reconstruction.tex
\begin{table}[!htbp]
\centering
\caption{\textbf{Dense reconstruction quality on Oxford Spires.}
CD denotes Chamfer distance.
Best/second-best: \textbf{bold}/\underline{underline}.}
\label{tab:supp_oxford_dense_reconstruction}
\fontsize{8}{9.5}\selectfont
\setlength{\tabcolsep}{6pt}
\renewcommand{\arraystretch}{1.05}
\begin{tabular}{lrr}
\toprule
\textbf{Method} & \textbf{CD (m)}\,$\downarrow$ & \textbf{F1@4\,m (\%)}\,$\uparrow$ \\
\midrule
LingBot-Map~\cite{chen2026lingbot} & 1.477 & \underline{90.58} \\
HorizonStream + LC~\cite{cheng2026horizonstream} & 1.946 & 85.96 \\
ABot-Recon + LC~\cite{amap2026abot} & 1.556 & 89.98 \\
CLoSeR~\cite{li2026closer} & \underline{1.461} & 90.41 \\
VGGT-Long ($\pi^3$)~\cite{deng2026vggtlong,wang2025pi3} & 2.011 & 85.91 \\
\midrule
\rowcolor[rgb]{0.965,0.950,0.977}
\textbf{VGGT-Long++ (ours)} & \textbf{1.351} & \textbf{92.04} \\
\bottomrule
\end{tabular}
\end{table}

%% file: tables/supplement/geometry_consistency_ablation.tex
\begin{table}[!htbp]
\centering
\caption{\textbf{Geometric consistency ablations on KITTI.}
Mean depth-normalized 3D discrepancy (\%, $\downarrow$).
All variants use scale-wise graph optimization and loop closure.}
\label{tab:supp_geometry_consistency_ablation}
\fontsize{8}{9.5}\selectfont
\setlength{\tabcolsep}{7pt}
\renewcommand{\arraystretch}{1.05}
\begin{tabular}{cccr}
\toprule
\textbf{Adapt.} & \textbf{Depth} & \textbf{Affine} & \textbf{Mean disc.} \\
\midrule
\xmark & \xmark & \xmark & 5.42 \\
\cmark & \xmark & \xmark & 3.76 \\
\cmark & \cmark & \xmark & \underline{2.67} \\
\rowcolor[rgb]{0.965,0.950,0.977}
\cmark & \cmark & \cmark & \textbf{1.82} \\
\bottomrule
\end{tabular}
\end{table}

%% file: tables/supplement/indoor_geometry_ablation.tex
\begin{table}[!htbp]
\centering
\caption{\textbf{Reconstruction quality ablations on NeuralRGBD.}
Acc./Comp. (cm, $\downarrow$), F1 at 5\,cm (\%, $\uparrow$).}
\label{tab:supp_indoor_geometry_ablation}
\fontsize{7.3}{9}\selectfont
\setlength{\tabcolsep}{1.1pt}
\renewcommand{\arraystretch}{1.02}
\begin{tabular*}{\columnwidth}{@{\extracolsep{\fill}}lrrr@{\hspace{4pt}}rrr@{\hspace{4pt}}rrr@{}}
\toprule
\multirow{2}{*}{\textbf{Sequence}}
& \multicolumn{3}{c}{\textbf{--}}
& \multicolumn{3}{c}{\textbf{+ depth}}
& \multicolumn{3}{c}{\textbf{+ affine}} \\
\cmidrule(lr){2-4}\cmidrule(lr){5-7}\cmidrule(l){8-10}
& Acc. & Comp. & F1 & Acc. & Comp. & F1 & Acc. & Comp. & F1 \\
\midrule
Breakfast Room & \underline{4.54} & \textbf{1.97} & \textbf{81.30} & \textbf{4.53} & \textbf{1.97} & \underline{81.18} & 4.58 & \underline{2.06} & 80.36 \\
Complete Kitchen & 4.77 & 2.13 & 78.52 & \underline{4.74} & \underline{2.09} & \underline{78.87} & \textbf{4.64} & \textbf{2.04} & \textbf{79.77} \\
Green Room & 3.87 & \underline{7.37} & 85.21 & \textbf{3.82} & \textbf{7.35} & \textbf{85.77} & \underline{3.84} & 7.54 & \underline{85.36} \\
Grey White Room & \underline{3.29} & 1.12 & 89.26 & 3.30 & \underline{1.10} & \underline{89.84} & \textbf{3.24} & \textbf{1.01} & \textbf{90.38} \\
Kitchen & 8.53 & 3.27 & 69.77 & \underline{8.39} & \underline{3.11} & \underline{70.16} & \textbf{8.34} & \textbf{3.06} & \textbf{70.49} \\
Morning Apartment & 4.51 & \textbf{1.28} & \underline{89.07} & \underline{4.43} & \textbf{1.28} & 88.99 & \textbf{4.40} & \underline{1.31} & \textbf{89.22} \\
Staircase & 4.85 & 0.99 & 84.07 & \underline{4.75} & \underline{0.93} & \underline{85.12} & \textbf{4.68} & \textbf{0.89} & \textbf{86.08} \\
Thin Geometry & 4.38 & \textbf{1.02} & \underline{84.67} & \underline{4.32} & \underline{1.05} & 84.61 & \textbf{4.27} & 1.06 & \textbf{84.97} \\
White Room & 3.72 & 0.93 & \underline{88.62} & \underline{3.70} & \underline{0.87} & 88.53 & \textbf{3.49} & \textbf{0.80} & \textbf{89.53} \\
\midrule
\rowcolor[rgb]{0.965,0.950,0.977}
\textbf{Mean} & 4.72 & 2.23 & 83.39 & \underline{4.66} & \textbf{2.19} & \underline{83.68} & \textbf{4.61} & \underline{2.20} & \textbf{84.02} \\
\bottomrule
\end{tabular*}
\end{table}

%% file: sec/supplement/other_methods.tex
\section{Relation to other methods}
\label{sec:supp_other_methods}

\subsection{Comparison with LASER}
\label{sec:supp_comparison_laser}

Even after similarity alignment, depth discrepancies between
chunks can vary across image regions.
LASER~\cite{ding2026laser} estimates scale factors for matched depth
layers in the overlap between consecutive windows and propagates
them forward through the incoming window.
VGGT-Long++ constrains the relative adjustments of layers predicted
for the same image using the median log-depth difference over
their mask intersection.
With chunk scales fixed, these log-depth adjustments are jointly
estimated across local, keyframe, and loop chunks by minimizing
the regularized weighted least-squares objective in
Eq.~\mainequation{eq:depth_alignment}.
Constraints from newly integrated chunks can therefore refine
layer depths in earlier local chunks.

\subsection{Comparison with VGGT-SLAM 2.0}
\label{sec:supp_comparison_vggt_slam2}

VGGT-SLAM 2.0~\cite{maggio2026vggtslam2} reconstructs dense submaps
from selected keyframes and aligns them using a global
$\mathrm{SL}(4)$ factor graph.
Intrinsic differences for an image shared by two submaps determine
the affine component of the measurement linking the corresponding
keyframe nodes.
VGGT-Long++ uses these differences as relative calibration constraints
to refine overlapping local chunks that cover all input frames.
With their global $\mathrm{Sim}(3)$ transformations fixed,
it jointly estimates one four-parameter intrinsic offset per chunk
through regularized least squares.
For each local chunk, the resulting offset then defines a shared
affine transformation for its 3D points and camera centers.
Calibration constraints from overlapping images can therefore
refine the geometry of frames outside the overlap.

%% file: sec/supplement/reconstructions.tex
\clearpage

\begin{figure*}[p]
\centering
\includegraphics[width=\linewidth]{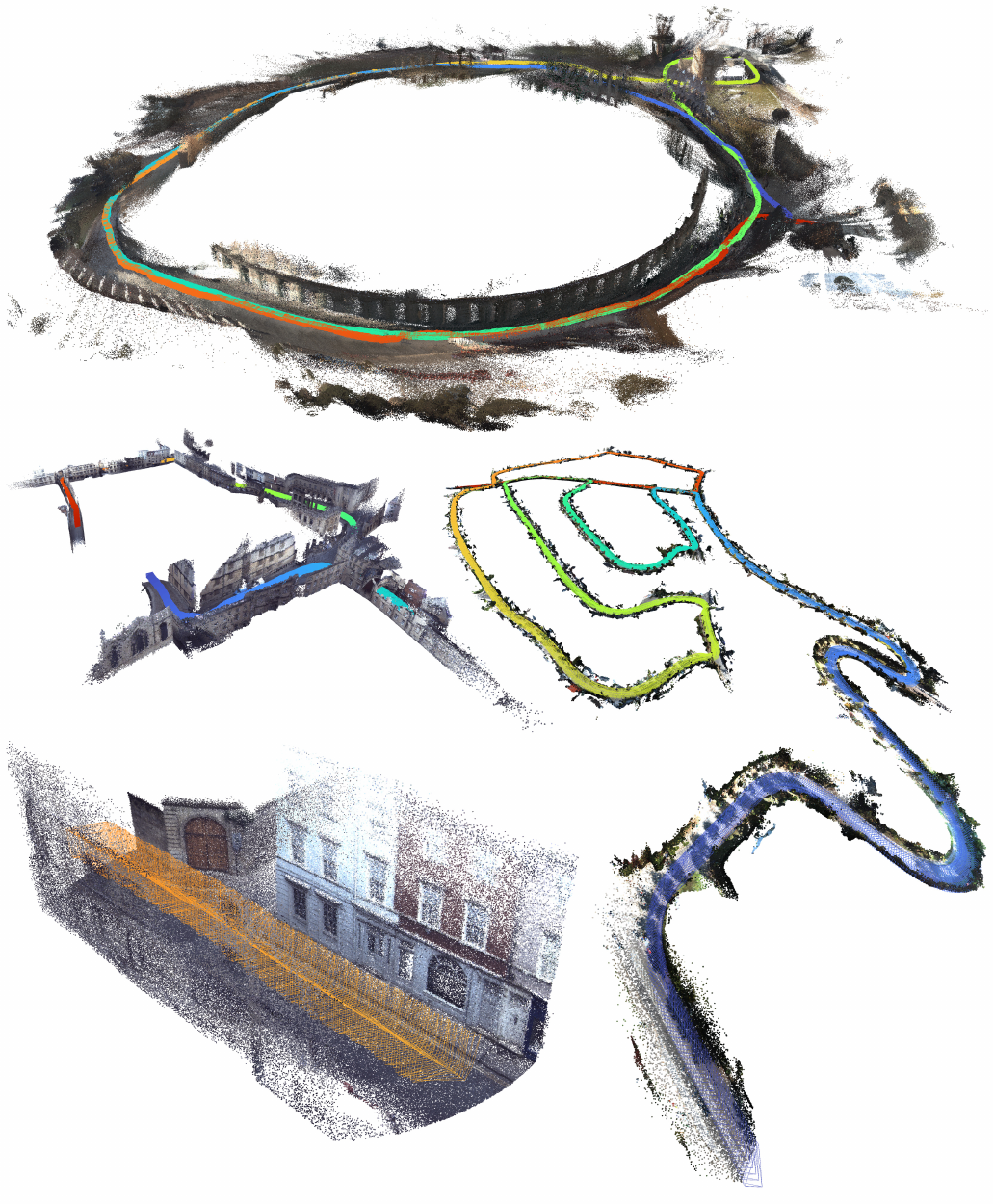}
\caption{\textbf{Long-sequence 3D reconstructions on VBR (top),
Oxford Spires (left), and KITTI (right).}}
\label{fig:supp_large_scale_reconstructions}
\end{figure*}

\begin{figure*}[p]
\centering
\includegraphics[width=\linewidth]{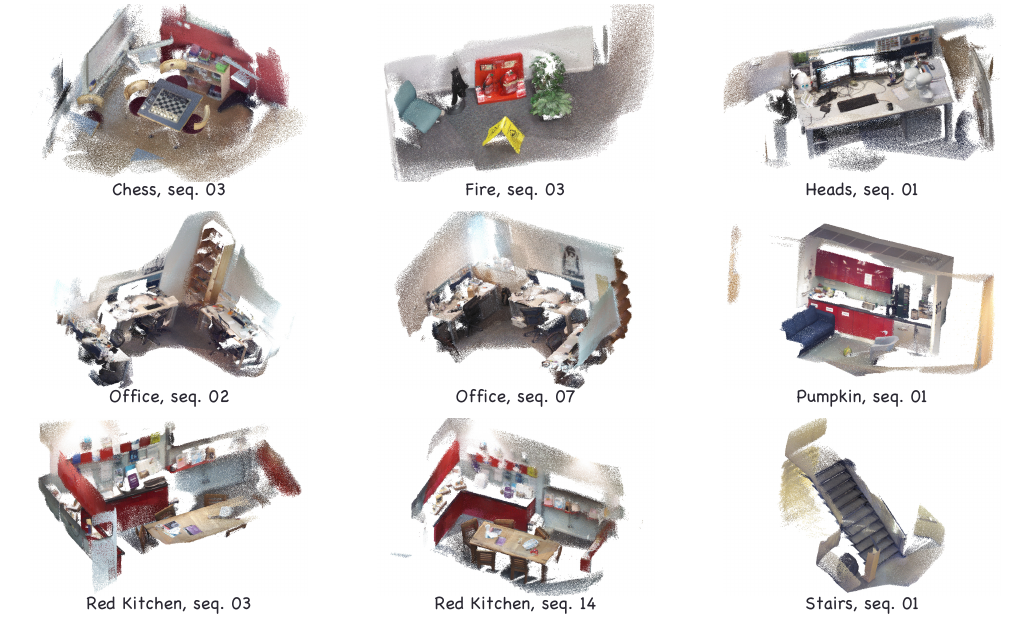}
\caption{\textbf{Reconstructions on 7-Scenes.}}
\label{fig:supp_7scenes_maps}
\end{figure*}

\begin{figure*}[p]
\centering
\includegraphics[width=\linewidth]{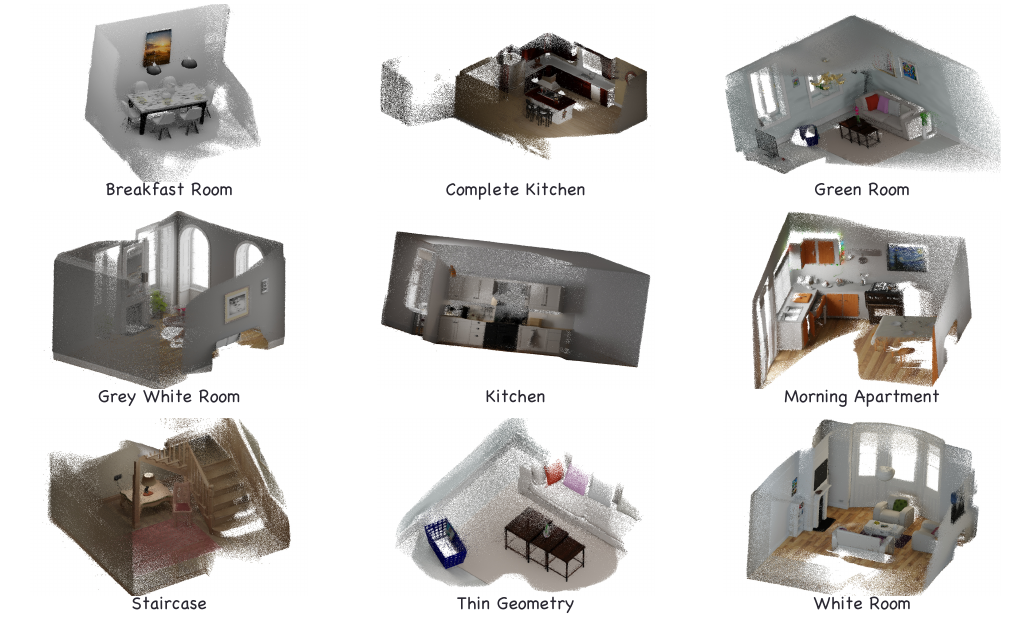}
\caption{\textbf{Reconstructions on NeuralRGBD.}}
\label{fig:supp_nrgbd_maps}
\end{figure*}

%% file: main.bib
@inproceedings{wang2024dust3r,
  author = {Shuzhe Wang and Vincent Leroy and Yohann Cabon and Boris Chidlovskii and J{\'e}r{\^o}me Revaud},
  title = {{DUSt3R: Geometric 3D Vision Made Easy}},
  booktitle = {Proceedings of the IEEE/CVF Conference on Computer Vision and Pattern Recognition},
  pages = {20697--20709},
  publisher = {IEEE},
  year = {2024},
  doi = {10.1109/cvpr52733.2024.01956},
  url = {https://openaccess.thecvf.com/content/CVPR2024/html/Wang_DUSt3R_Geometric_3D_Vision_Made_Easy_CVPR_2024_paper.html},
  isbn = {9798350353006},
  issn = {1063-6919}
}

@inproceedings{wang2025vggt,
  author = {Jianyuan Wang and Minghao Chen and Nikita Karaev and Andrea Vedaldi and Christian Rupprecht and David Novotny},
  title = {{VGGT: Visual Geometry Grounded Transformer}},
  booktitle = {Proceedings of the IEEE/CVF Conference on Computer Vision and Pattern Recognition},
  pages = {5294--5306},
  publisher = {IEEE},
  year = {2025},
  doi = {10.1109/cvpr52734.2025.00499},
  url = {https://openaccess.thecvf.com/content/CVPR2025/html/Wang_VGGT_Visual_Geometry_Grounded_Transformer_CVPR_2025_paper.html},
  isbn = {9798331543648},
  issn = {1063-6919}
}

@inproceedings{wang2025pi3,
  author = {Yifan Wang and Jianjun Zhou and Haoyi Zhu and Wenzheng Chang and Yang Zhou and Zizun Li and Junyi Chen and Jiangmiao Pang and Chunhua Shen and Tong He},
  title = {{$\pi^3$: Permutation-Equivariant Visual Geometry Learning}},
  booktitle = {International Conference on Learning Representations},
  pages = {10481--10497},
  year = {2026},
  url = {https://proceedings.iclr.cc/paper_files/paper/2026/hash/11a09e0aaa74867c6b0719c639fc09f8-Abstract-Conference.html}
}

@inproceedings{deng2026vggtlong,
  author = {Kai Deng and Zexin Ti and Jiawei Xu and Jian Yang and Jin Xie},
  title = {{VGGT-Long: Chunk it, Loop it, Align it -- Pushing VGGT's Limits on Kilometer-scale Long RGB Sequences}},
  booktitle = {IEEE International Conference on Robotics and Automation},
  pages = {5446--5453},
  publisher = {IEEE},
  year = {2026},
  doi = {10.1109/icra57385.2026.11697137},
  url = {https://doi.org/10.1109/icra57385.2026.11697137}
}

@inproceedings{maggio2025vggtslam,
  author = {Dominic Maggio and Hyungtae Lim and Luca Carlone},
  title = {{VGGT-SLAM: Dense RGB SLAM Optimized on the $\mathrm{SL}(4)$ Manifold}},
  booktitle = {Advances in Neural Information Processing Systems},
  volume = {38},
  pages = {129839--129867},
  publisher = {Curran Associates, Inc.},
  year = {2025},
  doi = {10.52202/085713-4324},
  url = {https://proceedings.neurips.cc/paper_files/paper/2025/hash/bc65ab11abfbad890171109686233f4e-Abstract-Conference.html}
}

@inproceedings{maggio2026vggtslam2,
  author = {Dominic Rosario Maggio and Luca Carlone},
  title = {{VGGT-SLAM 2.0: Real-time Dense Feed-forward Scene Reconstruction}},
  booktitle = {Proceedings of Robotics: Science and Systems},
  volume = {22},
  eid = {051},
  publisher = {Robotics: Science and Systems Foundation},
  year = {2026},
  doi = {10.15607/RSS.2026.XXII.051},
  url = {https://www.roboticsproceedings.org/rss22/p051.html},
  isbn = {979-8-9902848-2-1}
}

@inproceedings{ding2026laser,
  author = {Tianye Ding and Yiming Xie and Yiqing Liang and Moitreya Chatterjee and Pedro Miraldo and Huaizu Jiang},
  title = {{LASER: Layer-wise Scale Alignment for Training-Free Streaming 4D Reconstruction}},
  booktitle = {Proceedings of the IEEE/CVF Conference on Computer Vision and Pattern Recognition},
  pages = {36433--36443},
  year = {2026},
  url = {https://openaccess.thecvf.com/content/CVPR2026/html/Ding_LASER_Layer-wise_Scale_Alignment_for_Training-Free_Streaming_4D_Reconstruction_CVPR_2026_paper.html}
}

@article{jiang2026geoweaver,
  author = {Tinghao Jiang and Sheng Tang and Shengzhe Wei and Juntong Fang and Weiqi Zhang and Junsheng Zhou and Zesong Li},
  title = {{GeoWeaver: Accurate Long-Sequence 3D Reconstruction via Hierarchical Geometric Assembly}},
  journal = {arXiv preprint arXiv:2608.17389},
  publisher = {arXiv},
  year = {2026},
  doi = {10.48550/arxiv.2608.17389},
  url = {https://arxiv.org/abs/2608.17389}
}

@inproceedings{zhou2026online3r,
  author = {Shunkai Zhou and Zike Yan and Fei Xue and Dong Wu and Yuchen Deng and Hongbin Zha},
  title = {{Online3R: Online Learning for Consistent Sequential Reconstruction Based on Geometry Foundation Model}},
  booktitle = {Proceedings of the IEEE/CVF Conference on Computer Vision and Pattern Recognition},
  pages = {36529--36538},
  year = {2026},
  url = {https://openaccess.thecvf.com/content/CVPR2026/html/Zhou_Online3R_Online_Learning_for_Consistent_Sequential_Reconstruction_Based_on_Geometry_CVPR_2026_paper.html}
}

@article{youn2026selfgeometry,
  author = {Seokhyun Youn and Dahyeon Kye and Sung-Ho Bae and Jihyong Oh},
  title = {{Self-Geometry: GT-Free and Plug-and-Play Test-Time Adaptation for Geometrically Consistent 3D Vision Foundation Models}},
  journal = {arXiv preprint arXiv:2608.10708},
  publisher = {arXiv},
  year = {2026},
  doi = {10.48550/arxiv.2608.10708},
  url = {https://arxiv.org/abs/2608.10708}
}

@inproceedings{geiger2012kitti,
  author = {Andreas Geiger and Philip Lenz and Raquel Urtasun},
  title = {Are We Ready for Autonomous Driving? The {KITTI} Vision Benchmark Suite},
  booktitle = {Proceedings of the IEEE Conference on Computer Vision and Pattern Recognition},
  pages = {3354--3361},
  publisher = {IEEE},
  year = {2012},
  doi = {10.1109/CVPR.2012.6248074},
  url = {https://www.cvlibs.net/datasets/kitti/}
}

@article{tao2025spires,
  author = {Tao, Yifu and Mu{\~n}oz-Ba{\~n}{\'o}n, Miguel {\'A}ngel and Zhang, Lintong and Wang, Jiahao and Fu, Lanke Frank Tarimo and Fallon, Maurice},
  title = {The {Oxford Spires} Dataset: Benchmarking Large-Scale {LiDAR}-Visual Localisation, Reconstruction and Radiance Field Methods},
  journal = {The International Journal of Robotics Research},
  volume = {45},
  number = {6},
  pages = {839--857},
  publisher = {SAGE Publications},
  year = {2026},
  doi = {10.1177/02783649251369905},
  url = {https://doi.org/10.1177/02783649251369905},
  issn = {0278-3649, 1741-3176}
}

@inproceedings{brizi2024vbr,
  author = {Brizi, Leonardo and Giacomini, Emanuele and Di Giammarino, Luca and Ferrari, Simone and Salem, Omar and De Rebotti, Lorenzo and Grisetti, Giorgio},
  title = {{VBR}: A Vision Benchmark in {Rome}},
  booktitle = {IEEE International Conference on Robotics and Automation},
  pages = {15868--15874},
  publisher = {IEEE},
  year = {2024},
  doi = {10.1109/ICRA57147.2024.10611395},
  url = {https://arxiv.org/abs/2404.11322},
  isbn = {9798350384574}
}

@inproceedings{leroy2024mast3r,
  author = {Vincent Leroy and Yohann Cabon and J{\'e}r{\^o}me Revaud},
  title = {{Grounding Image Matching in 3D with MASt3R}},
  booktitle = {Computer Vision -- ECCV 2024},
  series = {Lecture Notes in Computer Science},
  volume = {15130},
  pages = {71--91},
  publisher = {Springer Nature Switzerland},
  year = {2025},
  doi = {10.1007/978-3-031-73220-1_5},
  url = {https://doi.org/10.1007/978-3-031-73220-1_5},
  isbn = {9783031732195, 9783031732201},
  issn = {0302-9743, 1611-3349}
}

@inproceedings{lin2025depthanything3,
  author = {Haotong Lin and Sili Chen and Jun Hao Liew and Donny Y. Chen and Zhenyu Li and Yang Zhao and Sida Peng and Hengkai Guo and Xiaowei Zhou and Guang Shi and Jiashi Feng and Bingyi Kang},
  title = {{Depth Anything 3: Recovering the Visual Space from Any Views}},
  booktitle = {International Conference on Learning Representations},
  pages = {141261--141285},
  year = {2026},
  url = {https://proceedings.iclr.cc/paper_files/paper/2026/hash/e4cd50120b6d7e8daff1749d6bbaa889-Abstract-Conference.html}
}

@inproceedings{keetha2026mapanything,
  author = {Nikhil Keetha and Norman M{\"u}ller and Johannes Sch{\"o}nberger and Lorenzo Porzi and Yuchen Zhang and Tobias Fischer and Arno Knapitsch and Duncan Zauss and Ethan Weber and Nelson Antunes and Jonathon Luiten and Manuel Lopez-Antequera and Samuel {Rota Bul{\`o}} and Christian Richardt and Deva Ramanan and Sebastian Scherer and Peter Kontschieder},
  title = {{MapAnything: Universal Feed-Forward Metric 3D Reconstruction}},
  booktitle = {International Conference on 3D Vision},
  pages = {499--509},
  publisher = {IEEE},
  year = {2026},
  doi = {10.1109/3dv69130.2026.00054},
  url = {https://doi.org/10.1109/3dv69130.2026.00054},
  isbn = {9798331573126},
  issn = {2378-3826}
}

@inproceedings{wang2025continuous,
  author = {Qianqian Wang and Yifei Zhang and Aleksander Holynski and Alexei A. Efros and Angjoo Kanazawa},
  title = {{Continuous 3D Perception Model with Persistent State}},
  booktitle = {Proceedings of the IEEE/CVF Conference on Computer Vision and Pattern Recognition},
  pages = {10510--10522},
  publisher = {IEEE},
  year = {2025},
  doi = {10.1109/cvpr52734.2025.00983},
  url = {https://openaccess.thecvf.com/content/CVPR2025/html/Wang_Continuous_3D_Perception_Model_with_Persistent_State_CVPR_2025_paper.html},
  isbn = {9798331543648},
  issn = {1063-6919}
}

@inproceedings{chen2025long3r,
  author = {Zhuoguang Chen and Minghui Qin and Tianyuan Yuan and Zhe Liu and Hang Zhao},
  title = {{LONG3R: Long Sequence Streaming 3D Reconstruction}},
  booktitle = {Proceedings of the IEEE/CVF International Conference on Computer Vision},
  pages = {5273--5284},
  publisher = {IEEE},
  year = {2025},
  doi = {10.1109/iccv51701.2025.00501},
  url = {https://openaccess.thecvf.com/content/ICCV2025/html/Chen_LONG3R_Long_Sequence_Streaming_3D_Reconstruction_ICCV_2025_paper.html}
}

@inproceedings{zhuo2025streamvggt,
  author = {Dong Zhuo and Wenzhao Zheng and Jiahe Guo and Yuqi Wu and Jie Zhou and Jiwen Lu},
  title = {{Streaming Visual Geometry Transformer}},
  booktitle = {International Conference on Learning Representations},
  pages = {88055--88072},
  year = {2026},
  url = {https://proceedings.iclr.cc/paper_files/paper/2026/hash/8da04a60948be713dc766f0c7e3a5b1f-Abstract-Conference.html}
}

@inproceedings{zhang2026loger,
  author = {Junyi Zhang and Charles Herrmann and Junhwa Hur and Chen Sun and Ming-Hsuan Yang and Forrester Cole and Trevor Darrell and Deqing Sun},
  title = {{LoGeR: Long-Context Geometric Reconstruction with Hybrid Memory}},
  booktitle = {Computer Vision -- ECCV 2026},
  series = {Lecture Notes in Computer Science},
  volume = {17072},
  pages = {263--281},
  publisher = {Springer Nature Switzerland},
  year = {2026},
  doi = {10.1007/978-3-032-37135-5_15},
  url = {https://doi.org/10.1007/978-3-032-37135-5_15},
  isbn = {9783032371348, 9783032371355},
  issn = {0302-9743, 1611-3349}
}

@article{liu2026suremap,
  author = {Mingkai Liu and Hao Zhao and Xingxing Zuo},
  title = {{SURE-Map: Self-Correcting Streaming Geometric Foundation Models}},
  journal = {arXiv preprint arXiv:2609.15795},
  publisher = {arXiv},
  year = {2026},
  doi = {10.48550/arxiv.2609.15795},
  url = {https://arxiv.org/abs/2609.15795}
}

@article{choi2026vggtbridge,
  author = {Sungjae Choi and Hanna Bae and Sunghyun Baek and Junmo Kim},
  title = {{VGGT-Bridge: Beyond Sequential Pose Graphs via Coarse-Stride Skip Edges}},
  journal = {arXiv preprint arXiv:2610.06594},
  publisher = {arXiv},
  year = {2026},
  doi = {10.48550/arxiv.2610.06594},
  url = {https://arxiv.org/abs/2610.06594}
}

@article{wang2026amb3rslam,
  author = {Hengyi Wang and Lourdes Agapito},
  title = {{AMB3R-SLAM: Kilometer-scale SLAM with Hierarchical Backend}},
  journal = {arXiv preprint arXiv:2609.19518},
  publisher = {arXiv},
  year = {2026},
  doi = {10.48550/arxiv.2609.19518},
  url = {https://arxiv.org/abs/2609.19518}
}

@inproceedings{hu2022lora,
  author = {Edward J. Hu and Yelong Shen and Phillip Wallis and Zeyuan Allen-Zhu and Yuanzhi Li and Shean Wang and Lu Wang and Weizhu Chen},
  title = {{LoRA: Low-Rank Adaptation of Large Language Models}},
  booktitle = {International Conference on Learning Representations},
  year = {2022},
  url = {https://openreview.net/forum?id=nZeVKeeFYf9}
}

@inproceedings{chen2026ttt3r,
  author = {Xingyu Chen and Yue Chen and Yuliang Xiu and Andreas Geiger and Anpei Chen},
  title = {{TTT3R: 3D Reconstruction as Test-Time Training}},
  booktitle = {International Conference on Learning Representations},
  pages = {50694--50718},
  year = {2026},
  url = {https://proceedings.iclr.cc/paper_files/paper/2026/hash/530eb593743559fa3c926a1fc1f520e9-Abstract-Conference.html}
}

@inproceedings{xie2026scal3r,
  author = {Tao Xie and Peishan Yang and Yudong Jin and Yingfeng Cai and Wei Yin and Weiqiang Ren and Qian Zhang and Wei Hua and Sida Peng and Xiaoyang Guo and Xiaowei Zhou},
  title = {{Scal3R: Scalable Test-Time Training for Large-Scale 3D Reconstruction}},
  booktitle = {Proceedings of the IEEE/CVF Conference on Computer Vision and Pattern Recognition},
  pages = {21760--21771},
  year = {2026},
  url = {https://openaccess.thecvf.com/content/CVPR2026/html/Xie_Scal3R_Scalable_Test-Time_Training_for_Large-Scale_3D_Reconstruction_CVPR_2026_paper.html}
}

@inproceedings{zhang2026vggtalign,
  author = {Wei Zhang and Yihang Wu and Songhua Li and Qi Wang},
  title = {{VGGT-Align: Bridging Local Reconstruction and Global Consistency for Long-Sequence 3D Reconstruction}},
  booktitle = {Proceedings of the 34th ACM International Conference on Multimedia},
  pages = {14830--14839},
  publisher = {ACM},
  year = {2026},
  doi = {10.1145/3767308.3836543},
  url = {https://doi.org/10.1145/3767308.3836543},
  isbn = {9798400722134}
}

@inproceedings{schoenberger2016sfm,
  author = {Johannes L. Sch{\"o}nberger and Jan-Michael Frahm},
  title = {{Structure-from-Motion Revisited}},
  booktitle = {Proceedings of the IEEE Conference on Computer Vision and Pattern Recognition},
  pages = {4104--4113},
  publisher = {IEEE},
  year = {2016},
  doi = {10.1109/cvpr.2016.445},
  url = {https://openaccess.thecvf.com/content_cvpr_2016/html/Schonberger_Structure-From-Motion_Revisited_CVPR_2016_paper.html},
  isbn = {9781467388511},
  issn = {1063-6919}
}

@article{campos2021orbslam3,
  author = {Carlos Campos and Richard Elvira and Juan J. G{\'o}mez Rodr{\'i}guez and Jos{\'e} M. M. Montiel and Juan D. Tard{\'o}s},
  title = {{ORB-SLAM3: An Accurate Open-Source Library for Visual, Visual-Inertial, and Multimap SLAM}},
  journal = {IEEE Transactions on Robotics},
  volume = {37},
  number = {6},
  pages = {1874--1890},
  publisher = {Institute of Electrical and Electronics Engineers (IEEE)},
  year = {2021},
  doi = {10.1109/TRO.2021.3075644},
  url = {https://arxiv.org/abs/2007.11898},
  issn = {1552-3098, 1941-0468}
}

@inproceedings{yang2025fast3r,
  author = {Jianing Yang and Alexander Sax and Kevin J. Liang and Mikael Henaff and Hao Tang and Ang Cao and Joyce Chai and Franziska Meier and Matt Feiszli},
  title = {{Fast3R: Towards 3D Reconstruction of 1000+ Images in One Forward Pass}},
  booktitle = {Proceedings of the IEEE/CVF Conference on Computer Vision and Pattern Recognition},
  pages = {21924--21935},
  publisher = {IEEE},
  year = {2025},
  doi = {10.1109/cvpr52734.2025.02042},
  url = {https://openaccess.thecvf.com/content/CVPR2025/html/Yang_Fast3R_Towards_3D_Reconstruction_of_1000_Images_in_One_Forward_CVPR_2025_paper.html},
  isbn = {9798331543648},
  issn = {1063-6919}
}

@inproceedings{cabon2025must3r,
  author = {Yohann Cabon and Lucas Stoffl and Leonid Antsfeld and Gabriela Csurka and Boris Chidlovskii and J{\'e}r{\^o}me Revaud and Vincent Leroy},
  title = {{MUSt3R: Multi-view Network for Stereo 3D Reconstruction}},
  booktitle = {Proceedings of the IEEE/CVF Conference on Computer Vision and Pattern Recognition},
  pages = {1050--1060},
  publisher = {IEEE},
  year = {2025},
  doi = {10.1109/cvpr52734.2025.00106},
  url = {https://openaccess.thecvf.com/content/CVPR2025/html/Cabon_MUSt3R_Multi-view_Network_for_Stereo_3D_Reconstruction_CVPR_2025_paper.html},
  isbn = {9798331543648},
  issn = {1063-6919}
}

@inproceedings{wang2025spann3r,
  author = {Hengyi Wang and Lourdes Agapito},
  title = {{3D Reconstruction with Spatial Memory}},
  booktitle = {International Conference on 3D Vision},
  pages = {78--89},
  publisher = {IEEE},
  year = {2025},
  doi = {10.1109/3dv66043.2025.00013},
  url = {https://doi.org/10.1109/3dv66043.2025.00013},
  isbn = {9798331538514},
  issn = {2378-3826}
}

@article{yuan2026infinitevggt,
  author = {Shuai Yuan and Yantai Yang and Xiaotian Yang and Xupeng Zhang and Zhonghao Zhao and Lingming Zhang and Zhipeng Zhang},
  title = {{InfiniteVGGT: Visual Geometry Grounded Transformer for Endless Streams}},
  journal = {arXiv preprint arXiv:2601.02281},
  publisher = {arXiv},
  year = {2026},
  doi = {10.48550/arxiv.2601.02281},
  url = {https://arxiv.org/abs/2601.02281}
}

@inproceedings{cheng2026horizonstream,
  author = {Chong Cheng and Peilin Tao and Nanjie Yao and Guanzhi Ding and Xianda Chen and Yuansen Du and Xiaoyang Guo and Wei Yin and Weiqiang Ren and Qian Zhang and Zhengqing Chen and Hao Wang},
  title = {{HorizonStream: Long-Horizon Attention for Streaming 3D Reconstruction}},
  booktitle = {Advances in Neural Information Processing Systems},
  year = {2026},
  url = {https://arxiv.org/abs/2605.23889},
  note = {To appear}
}

@article{liu2026mem3r,
  author = {Changkun Liu and Jiezhi Yang and Zeman Li and Yuan Deng and Jiancong Guo and Binbin Xu and Luca Ballan},
  title = {{Mem3R: Streaming 3D Reconstruction with Hybrid Memory}},
  journal = {arXiv preprint arXiv:2604.07279},
  publisher = {arXiv},
  year = {2026},
  doi = {10.48550/arxiv.2604.07279},
  url = {https://arxiv.org/abs/2604.07279v2},
  eprint = {2604.07279},
  archiveprefix = {arXiv},
  version = {2}
}

@inproceedings{liu2025slam3r,
  author = {Yuzheng Liu and Siyan Dong and Shuzhe Wang and Yingda Yin and Yanchao Yang and Qingnan Fan and Baoquan Chen},
  title = {{SLAM3R: Real-Time Dense Scene Reconstruction from Monocular RGB Videos}},
  booktitle = {Proceedings of the IEEE/CVF Conference on Computer Vision and Pattern Recognition},
  pages = {16651--16662},
  publisher = {IEEE},
  year = {2025},
  doi = {10.1109/cvpr52734.2025.01552},
  url = {https://openaccess.thecvf.com/content/CVPR2025/html/Liu_SLAM3R_Real-Time_Dense_Scene_Reconstruction_from_Monocular_RGB_Videos_CVPR_2025_paper.html},
  isbn = {9798331543648},
  issn = {1063-6919}
}

@inproceedings{tao2026anchor3r,
  author = {Peilin Tao and Chong Cheng and Yuansen Du and Caiwei Song and Zhengqing Chen and Xiaoyang Guo and Wei Yin and Weiqiang Ren and Qian Zhang and Hainan Cui and Shuhan Shen},
  title = {{Anchor3R: Streaming 3D Reconstruction with Transient Anchors for Long-Horizon Visual Mapping}},
  booktitle = {Conference on Robot Learning},
  year = {2026},
  url = {https://openreview.net/forum?id=Q53ayGo2IE}
}

@article{li2026closer,
  author = {Moyang Li and Zihan Zhu and Wei Zhang and Marc Pollefeys and Daniel Barath},
  title = {{CLoSeR: Closing the Loop for Long-Context Streaming Reconstruction}},
  journal = {arXiv preprint arXiv:2610.01927},
  publisher = {arXiv},
  year = {2026},
  doi = {10.48550/arxiv.2610.01927},
  url = {https://arxiv.org/abs/2610.01927}
}

@inproceedings{sun2020ttt,
  author = {Yu Sun and Xiaolong Wang and Zhuang Liu and John Miller and Alexei A. Efros and Moritz Hardt},
  title = {{Test-Time Training with Self-Supervision for Generalization under Distribution Shifts}},
  booktitle = {Proceedings of the 37th International Conference on Machine Learning},
  series = {Proceedings of Machine Learning Research},
  volume = {119},
  pages = {9229--9248},
  publisher = {PMLR},
  year = {2020},
  url = {https://proceedings.mlr.press/v119/sun20b.html},
  issn = {2640-3498}
}

@inproceedings{wang2021tent,
  author = {Dequan Wang and Evan Shelhamer and Shaoteng Liu and Bruno Olshausen and Trevor Darrell},
  title = {{Tent: Fully Test-Time Adaptation by Entropy Minimization}},
  booktitle = {International Conference on Learning Representations},
  year = {2021},
  url = {https://openreview.net/forum?id=uXl3bZLkr3c}
}

@inproceedings{wang2022cotta,
  author = {Qin Wang and Olga Fink and Luc {Van Gool} and Dengxin Dai},
  title = {{Continual Test-Time Domain Adaptation}},
  booktitle = {Proceedings of the IEEE/CVF Conference on Computer Vision and Pattern Recognition},
  pages = {7201--7211},
  publisher = {IEEE},
  year = {2022},
  doi = {10.1109/cvpr52688.2022.00706},
  url = {https://openaccess.thecvf.com/content/CVPR2022/html/Wang_Continual_Test-Time_Domain_Adaptation_CVPR_2022_paper.html},
  isbn = {9781665469463},
  issn = {1063-6919}
}

@inproceedings{zhou2017sfmlearner,
  author = {Tinghui Zhou and Matthew Brown and Noah Snavely and David G. Lowe},
  title = {{Unsupervised Learning of Depth and Ego-Motion from Video}},
  booktitle = {Proceedings of the IEEE Conference on Computer Vision and Pattern Recognition},
  pages = {1851--1858},
  publisher = {IEEE},
  year = {2017},
  doi = {10.1109/cvpr.2017.700},
  url = {https://openaccess.thecvf.com/content_cvpr_2017/html/Zhou_Unsupervised_Learning_of_CVPR_2017_paper.html},
  isbn = {9781538604571},
  issn = {1063-6919}
}

@inproceedings{mahjourian2018geometric,
  author = {Reza Mahjourian and Martin Wicke and Anelia Angelova},
  title = {{Unsupervised Learning of Depth and Ego-Motion from Monocular Video Using 3D Geometric Constraints}},
  booktitle = {Proceedings of the IEEE/CVF Conference on Computer Vision and Pattern Recognition},
  pages = {5667--5675},
  publisher = {IEEE},
  year = {2018},
  doi = {10.1109/cvpr.2018.00594},
  url = {https://openaccess.thecvf.com/content_cvpr_2018/html/Mahjourian_Unsupervised_Learning_of_CVPR_2018_paper.html},
  isbn = {9781538664209},
  issn = {1063-6919}
}

@inproceedings{bian2019scaleconsistent,
  author = {Jia-Wang Bian and Zhichao Li and Naiyan Wang and Huangying Zhan and Chunhua Shen and Ming-Ming Cheng and Ian Reid},
  title = {{Unsupervised Scale-consistent Depth and Ego-motion Learning from Monocular Video}},
  booktitle = {Advances in Neural Information Processing Systems},
  volume = {32},
  pages = {35--45},
  publisher = {Curran Associates, Inc.},
  year = {2019},
  url = {https://proceedings.neurips.cc/paper_files/paper/2019/hash/6364d3f0f495b6ab9dcf8d3b5c6e0b01-Abstract.html}
}

@article{luo2020consistent,
  author = {Xuan Luo and Jia-Bin Huang and Richard Szeliski and Kevin Matzen and Johannes Kopf},
  title = {{Consistent Video Depth Estimation}},
  journal = {ACM Transactions on Graphics},
  volume = {39},
  number = {4},
  pages = {71:1--71:13},
  eid = {71},
  publisher = {Association for Computing Machinery (ACM)},
  year = {2020},
  doi = {10.1145/3386569.3392377},
  url = {https://roxanneluo.github.io/Consistent-Video-Depth-Estimation/},
  issn = {0730-0301, 1557-7368}
}

@article{felzenszwalb2004segmentation,
  author = {Pedro F. Felzenszwalb and Daniel P. Huttenlocher},
  title = {Efficient Graph-Based Image Segmentation},
  journal = {International Journal of Computer Vision},
  volume = {59},
  number = {2},
  pages = {167--181},
  publisher = {Kluwer Academic Publishers},
  year = {2004},
  doi = {10.1023/B:VISI.0000022288.19776.77},
  url = {https://cs.brown.edu/people/pfelzens/papers/seg-ijcv.pdf},
  issn = {0920-5691, 1573-1405}
}

@article{lowe2004sift,
  author = {David G. Lowe},
  title = {Distinctive Image Features from Scale-Invariant Keypoints},
  journal = {International Journal of Computer Vision},
  volume = {60},
  number = {2},
  pages = {91--110},
  publisher = {Kluwer Academic Publishers},
  year = {2004},
  doi = {10.1023/B:VISI.0000029664.99615.94},
  url = {https://www.cs.ubc.ca/~lowe/papers/ijcv04.pdf},
  issn = {0920-5691, 1573-1405}
}

@article{fischler1981ransac,
  author = {Martin A. Fischler and Robert C. Bolles},
  title = {{Random Sample Consensus: A Paradigm for Model Fitting with Applications to Image Analysis and Automated Cartography}},
  journal = {Communications of the ACM},
  volume = {24},
  number = {6},
  pages = {381--395},
  publisher = {Association for Computing Machinery},
  month = jun,
  year = {1981},
  doi = {10.1145/358669.358692},
  url = {https://doi.org/10.1145/358669.358692},
  issn = {0001-0782}
}

@book{hartley2004multiple,
  author = {Richard Hartley and Andrew Zisserman},
  title = {Multiple View Geometry in Computer Vision},
  edition = {Second},
  publisher = {Cambridge University Press},
  year = {2004},
  doi = {10.1017/CBO9780511811685},
  url = {https://doi.org/10.1017/CBO9780511811685},
  isbn = {9780521540513, 9780511811685}
}

@article{amap2026abot,
  author = {Jiarong Han and Jincheng Xiong and Yuzhou Liu and Linzhe Shi and Changjie Wu and Ning Guo and Mu Xu and Hang Zhang and Ming Qian},
  title = {Revisiting Local Context for Long-Horizon Streaming {3D} Reconstruction},
  journal = {arXiv preprint arXiv:2608.27529},
  publisher = {arXiv},
  year = {2026},
  doi = {10.48550/arxiv.2608.27529},
  url = {https://arxiv.org/abs/2608.27529v1}
}

@inproceedings{chen2026lingbot,
  author = {Lin-Zhuo Chen and Jian Gao and Shangzhan Zhang and Yihang Chen and Nan Xue and Jianyuan Wang and Christian Rupprecht and Xun Cao and Xing Zhu and Yujun Shen and Yao Yao and Yinghao Xu},
  title = {{Geometric Context Transformer for Streaming 3D Reconstruction}},
  booktitle = {Computer Vision -- ECCV 2026},
  series = {Lecture Notes in Computer Science},
  volume = {17037},
  pages = {293--314},
  publisher = {Springer Nature Switzerland},
  year = {2026},
  doi = {10.1007/978-3-032-37467-7_17},
  url = {https://doi.org/10.1007/978-3-032-37467-7_17},
  isbn = {9783032374660, 9783032374677},
  issn = {0302-9743, 1611-3349}
}

@inproceedings{li2026grfrecon,
  author = {Enpeng Li and Yunzhou Zhang and Zhiyao Zhang and Dexuan Lyu and Chenyu Wang and Chiyuan Cui and Cheng Cheng},
  title = {{GRF-Recon: Global Ray-Field Optimization for Long-Sequence Feed-Forward Reconstruction}},
  booktitle = {Computer Vision -- ECCV 2026},
  series = {Lecture Notes in Computer Science},
  volume = {17073},
  pages = {62--78},
  publisher = {Springer Nature Switzerland},
  year = {2026},
  doi = {10.1007/978-3-032-37138-6_4},
  url = {https://doi.org/10.1007/978-3-032-37138-6_4},
  isbn = {9783032371379, 9783032371386},
  issn = {0302-9743, 1611-3349}
}

@inproceedings{teed2023dpvo,
  author = {Zachary Teed and Lahav Lipson and Jia Deng},
  title = {Deep Patch Visual Odometry},
  booktitle = {Advances in Neural Information Processing Systems},
  volume = {36},
  pages = {39033--39051},
  publisher = {Curran Associates, Inc.},
  year = {2023},
  doi = {10.52202/075280-1696},
  url = {https://proceedings.neurips.cc/paper_files/paper/2023/hash/7ac484b0f1a1719ad5be9aa8c8455fbb-Abstract-Conference.html}
}

@inproceedings{shotton2013scenecoordinate,
  author = {Jamie Shotton and Ben Glocker and Christopher Zach and Shahram Izadi and Antonio Criminisi and Andrew Fitzgibbon},
  title = {Scene Coordinate Regression Forests for Camera Relocalization in {RGB-D} Images},
  booktitle = {Proceedings of the IEEE Conference on Computer Vision and Pattern Recognition},
  pages = {2930--2937},
  publisher = {IEEE},
  year = {2013},
  doi = {10.1109/cvpr.2013.377},
  url = {https://www.microsoft.com/en-us/research/project/rgb-d-dataset-7-scenes/}
}

@inproceedings{azinovic2022neuralrgbd,
  author = {Dejan Azinovi{\'c} and Ricardo Martin-Brualla and Dan B Goldman and Matthias Nie{\ss}ner and Justus Thies},
  title = {Neural {RGB-D} Surface Reconstruction},
  booktitle = {Proceedings of the IEEE/CVF Conference on Computer Vision and Pattern Recognition},
  pages = {6290--6301},
  publisher = {IEEE},
  year = {2022},
  doi = {10.1109/cvpr52688.2022.00619},
  url = {https://openaccess.thecvf.com/content/CVPR2022/html/Azinovic_Neural_RGB-D_Surface_Reconstruction_CVPR_2022_paper.html},
  isbn = {9781665469463},
  issn = {1063-6919}
}

@article{deng2026glob3r,
  author = {Junyuan Deng and Heng Li and Kejie Qiu and Lingteng Qiu and Rui Peng and Weichao Shen and Weihao Yuan and Siyu Zhu and Zilong Dong and Ping Tan},
  title = {{Glob3R}: Global Structure-from-Motion with {3D} Foundation Models},
  journal = {arXiv preprint arXiv:2607.09225},
  publisher = {arXiv},
  year = {2026},
  doi = {10.48550/arxiv.2607.09225},
  url = {https://arxiv.org/abs/2607.09225}
}

@article{fang2026slamformer,
  author = {Zhijian Fang and Weicheng Zheng and Yijun Yuan and Weibang Wang and Zhuoguang Chen and Chang Sun and Junhao Huang and Kenan Li and Minghui Qin and Hang Zhao},
  title = {{SLAMFormer-$\infty$}: Infinite {SLAM} Transformer for Unbounded Frontend and Backend Processing},
  journal = {arXiv preprint arXiv:2608.03429},
  publisher = {arXiv},
  year = {2026},
  doi = {10.48550/arxiv.2608.03429},
  url = {https://arxiv.org/abs/2608.03429}
}

@inproceedings{lipson2024dpvslam,
  author = {Lahav Lipson and Zachary Teed and Jia Deng},
  title = {Deep Patch Visual {SLAM}},
  booktitle = {Computer Vision -- ECCV 2024},
  series = {Lecture Notes in Computer Science},
  volume = {15060},
  pages = {424--440},
  publisher = {Springer Nature Switzerland},
  year = {2025},
  doi = {10.1007/978-3-031-72627-9_24},
  url = {https://doi.org/10.1007/978-3-031-72627-9_24},
  isbn = {9783031726262, 9783031726279},
  issn = {0302-9743, 1611-3349}
}

@inproceedings{li2026droidw,
  author = {Moyang Li and Zihan Zhu and Marc Pollefeys and Daniel Barath},
  title = {{DROID-SLAM} in the Wild},
  booktitle = {Proceedings of the IEEE/CVF Conference on Computer Vision and Pattern Recognition},
  pages = {36498--36508},
  publisher = {IEEE},
  year = {2026},
  url = {https://openaccess.thecvf.com/content/CVPR2026/html/Li_DROID-SLAM_in_the_Wild_CVPR_2026_paper.html}
}

@inproceedings{loshchilov2019adamw,
  author = {Ilya Loshchilov and Frank Hutter},
  title = {Decoupled Weight Decay Regularization},
  booktitle = {International Conference on Learning Representations},
  year = {2019},
  url = {https://openreview.net/forum?id=Bkg6RiCqY7}
}
